\documentclass[journal]{IEEEtran}
\usepackage{amsmath,amssymb}
\usepackage{colortbl}
\usepackage{diagbox}
\usepackage{subeqnarray}
\usepackage{cases}
\usepackage{setspace}
\usepackage{tabularx}
\usepackage{booktabs}
\usepackage{multicol}
\usepackage{dcolumn}
\usepackage{graphicx}
\usepackage{subfigure}
\usepackage{float}
\usepackage{epsfig}
\usepackage{amsmath}
\usepackage{amssymb}
\usepackage{array}
\usepackage{epstopdf}
\usepackage{color}
\usepackage{xcolor}
\usepackage[nocompress]{cite}
\usepackage{psfrag}
\usepackage{cite}
\usepackage{multirow}
\usepackage{color}

\graphicspath{{figures/}}
\usepackage[bookmarks,colorlinks,linkcolor=red, anchorcolor=blue, citecolor=green]{hyperref}
\usepackage{url}

\begin{document}
%
\title{FFDNet: Toward a Fast and Flexible Solution for CNN based Image Denoising}

\author{Kai~Zhang,
        Wangmeng Zuo,~\IEEEmembership{Senior Member,~IEEE,}
        and~Lei Zhang,~\IEEEmembership{Fellow,~IEEE}

\thanks{This project is partially supported by the National Natural Scientific Foundation of China (NSFC) under Grant No. 61671182 and 61471146, and the HK RGC GRF grant (under no. PolyU 152124/15E).}
\thanks{K. Zhang is with the School of Computer Science and Technology, Harbin
Institute of Technology, Harbin 150001, China, and also with the Department
of Computing, The Hong Kong Polytechnic University, Hong Kong (e-mail: cskaizhang@gmail.com).}
\thanks{W. Zuo is with the School of Computer Science and
Technology, Harbin Institute of Technology, Harbin 150001, China (e-mail:
cswmzuo@gmail.com).}
\thanks{L. Zhang is with the Department of Computing, The Hong Kong
Polytechnic University, Hong Kong (e-mail: cslzhang@comp.polyu.edu.hk).}
}

%


\maketitle

\begin{abstract}

Due to the fast inference and good performance, discriminative learning methods have been widely studied in image denoising. However, these methods mostly learn a specific model for each noise level, and require multiple models for denoising images with different noise levels. They also lack flexibility to deal with spatially variant noise, limiting their applications in practical denoising. To address these issues, we present a fast and flexible denoising convolutional neural network, namely FFDNet, with a tunable noise level map as the input. The proposed FFDNet works on downsampled sub-images, achieving a good trade-off between inference speed and denoising performance. In contrast to the existing discriminative denoisers, FFDNet enjoys several desirable properties, including (i) the ability to handle a wide range of noise levels (i.e., [0, 75]) effectively with a single network, (ii) the ability to remove spatially variant noise by specifying a non-uniform noise level map, and (iii) faster speed than benchmark BM3D even on CPU without sacrificing denoising performance. Extensive experiments on synthetic and real noisy images are conducted to evaluate FFDNet in comparison with state-of-the-art denoisers. The results show that FFDNet is effective and efficient, making it highly attractive for practical denoising applications.

\end{abstract}


%
\IEEEpeerreviewmaketitle

\section{Introduction}

\IEEEPARstart{T}{he} importance of image denoising in low level vision can be revealed from many aspects.
First, noise corruption is inevitable during the image sensing process and it may heavily degrade the visual quality of acquired image. Removing noise from the observed image is an essential step in various image processing and computer vision tasks~\cite{andrews1977digital,chatterjee2010denoising}.
Second, from the Bayesian perspective, image denoising is an ideal test bed for evaluating image prior models and optimization methods~\cite{roth2005fields,zoran2011learning,gu2014weighted}.
Last but not least, in the unrolled inference via variable splitting techniques, many image restoration problems can be addressed by sequentially solving a series of denoising subproblems, which further broadens the application fields of image denoising~\cite{afonso2010fast,heide2014flexisp,romano2016little,zhang2017learning}.

As in many previous literature of image denoising~\cite{portilla2003image,dabov2007image,mairal2009non,dong2013nonlocally}, in this paper we assume that the noise is additive white Gaussian noise (AWGN) and the noise level is given. In order to handle practical image denoising problems, a flexible image denoiser is expected to have the following desirable properties: (i) it is able to perform denoising using a single model; (ii) it is efficient, effective and user-friendly; and (iii) it can handle spatially variant noise. Such a denoiser can be directly deployed to recover the clean image when the noise level is known or can be well estimated. When the noise level is unknown or is difficult to estimate, the denoiser should allow the user to adaptively control the trade-off between noise reduction and detail preservation. Furthermore, the noise can be spatially variant and the denoiser should be flexible enough to handle spatially variant noise.

However, state-of-the-art image denoising methods are still limited in flexibility or efficiency. In general, image denoising methods can be grouped into two major categories, model-based methods and discriminative learning based ones. Model-based methods such as BM3D~\cite{dabov2007image} and WNNM~\cite{gu2014weighted} are flexible in handling denoising problems with various noise levels, but they suffer from several drawbacks. For example, their optimization algorithms are generally time-consuming, and cannot be directly used to remove spatially variant noise. Moreover,
model-based methods usually employ hand-crafted image priors (e.g., sparsity~\cite{elad2006image,mairal2008sparse} and nonlocal self-similarity~\cite{buades2005non,mairal2009non,dong2013nonlocally}), which may not be strong enough to characterize complex image structures.

As an alternative, discriminative denoising methods aim to learn the underlying image prior and fast inference from a training set of degraded and ground-truth image pairs. One approach is to learn stage-wise image priors in the context of truncated inference procedure~\cite{chen2015trainable}.
Another more popular approach is plain discriminative learning, such as the MLP~\cite{burger2012image} and convolutional neural network (CNN) based methods~\cite{jain2009natural,zhang2017beyond}, among which the DnCNN~\cite{zhang2017beyond} method has achieved very competitive denoising performance. The success of CNN for image denoising is attributed to its large modeling capacity and tremendous advances in network training and design.
However, existing discriminative denoising methods are limited in flexibility, and the learned model is usually tailored to a specific noise level. From the perspective of regression, they aim to learn a mapping function
$\emph{\textbf{x}}=\mathcal{F}(\emph{\textbf{y}};\Theta_{\sigma})$ between the input noisy observation $\emph{\textbf{y}}$ and the desired output $\emph{\textbf{x}}$. The model parameters $\Theta_{\sigma}$ are trained for noisy images corrupted by AWGN with a fixed noise level $\sigma$, while the trained model with $\Theta_{\sigma}$ is hard to be directly deployed to images with other noise levels. Though a single CNN model (i.e., DnCNN-B) is trained in~\cite{zhang2017beyond} for Gaussian denoising, it does not generalize well to real noisy images and works only if the noise level is in the preset range, e.g., $[0, 55]$. Besides, all the existing discriminative learning based methods lack flexibility to deal with spatially variant noise.

To overcome the drawbacks of existing CNN based denoising methods, we present a fast and flexible denoising convolutional neural network (FFDNet). Specifically, our FFDNet is formulated as $\emph{\textbf{x}}=\mathcal{F}(\emph{\textbf{y}}, \emph{\textbf{M}};\Theta)$, where $\emph{\textbf{M}}$ is a noise level map. In the DnCNN model $\emph{\textbf{x}}=\mathcal{F}(\emph{\textbf{y}};\Theta_{\sigma})$, the parameters $\Theta_{\sigma}$ vary with the change of noise level $\sigma$, while in the FFDNet model, the noise level map is modeled as an input and the model parameters $\Theta$ are invariant to noise level. Thus, FFDNet provides a flexible way to handle different noise levels with a single network.

By introducing a noise level map as input, it is natural to expect that the model performs well when the noise level map matches the ground-truth one of noisy input.
Furthermore, the noise level map should also play the role of controlling the trade-off between noise reduction and detail preservation.
It is found that heavy visual quality degradation may be engendered when setting a larger noise level to smooth out the details.
We highlight this problem and adopt a method of orthogonal initialization on convolutional filters to alleviate this.
Besides, the proposed FFDNet works on downsampled sub-images, which largely accelerates the training and testing speed, and enlarges the receptive field as well.

Using images corrupted by AWGN, we quantitatively compare FFDNet with state-of-the-art denoising methods, including model-based methods such as BM3D~\cite{dabov2007image} and WNNM~\cite{gu2014weighted} and discriminative learning based methods such as TNRD~\cite{chen2015trainable} and DnCNN~\cite{zhang2017beyond}.
The results clearly demonstrate the superiority of FFDNet in terms of both denoising performance and computational efficiency. In addition, FFDNet performs favorably on images corrupted by spatially variant AWGN. We further evaluate FFDNet on real-world noisy images, where the noise is often signal-dependent, non-Gaussian and spatially variant. The proposed FFDNet model still achieves perceptually convincing results by setting proper noise level maps. Overall, FFDNet enjoys high potentials for practical denoising applications.

The main contribution of our work is summarized as follows:
\begin{itemize}

\item A fast and flexible denoising network, namely FFDNet, is proposed for discriminative image denoising. By taking a tunable noise level map as input, a single FFDNet is able to deal with noise on different levels, as well as spatially variant noise.

\item We highlight the importance to guarantee the role of the noise level map in controlling the trade-off between noise reduction and detail preservation.

\item FFDNet exhibits perceptually appealing results on both synthetic noisy images corrupted by AWGN and real-world noisy images, demonstrating its potential for practical image denoising.

\end{itemize}

The remainder of this paper is organized as follows.
Sec.~\ref{sec:related_work} reviews existing discriminative denoising methods.
Sec.~\ref{sec:method} presents the proposed image denoising model.
Sec.~\ref{sec:experiments} reports the experimental results.
Sec.~\ref{sec:conclusion} concludes the paper.

\section{Related Work}
\label{sec:related_work}

In this section, we briefly review and discuss the two major categories of relevant methods to this work, i.e., maximum a posteriori (MAP) inference guided discriminative learning and plain discriminative learning.

\vspace{-0.2cm}
\subsection{MAP Inference Guided Discriminative Learning}
Instead of first learning the prior and then performing the inference, this category of methods aims to learn the prior parameters along with a compact unrolled inference through minimizing a loss function~\cite{barbu2009training}.
Following the pioneer work of fields of experts~\cite{roth2005fields}, Barbu~\cite{barbu2009training} trained a discriminative Markov random field (MRF) model together with a gradient descent inference for image denoising. Samuel and Tappen~\cite{samuel2009learning} independently proposed a compact gradient descent inference learning framework, and discussed the advantages of discriminative learning over model-based optimization method with MRF prior. Sun and Tappen~\cite{sun2011learning} proposed a novel nonlocal range MRF (NLR-MRF) framework, and employed the gradient-based discriminative learning method to train the model. Generally speaking, the methods above only learn the prior parameters in a discriminative manner, while the inference parameters are stage-invariant.

\begin{figure*}[htbp]
  \centering
  \includegraphics[width=1\textwidth]{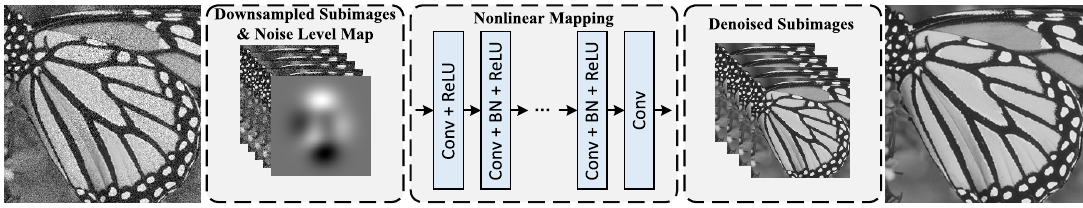}\\
  \caption{The architecture of the proposed FFDNet for image denoising. The input image is reshaped to four sub-images, which are then input to the CNN together with a noise level map. The final output is reconstructed by the four denoised sub-images.}
  \label{fig1}\vspace{-0.1cm}
\end{figure*}

With the aid of unrolled half quadratic splitting (HQS) techniques, Schmidt et al.~\cite{schmidt2014shrinkage,tuprints6044} proposed a cascade of shrinkage fields (CSF) framework to learn stage-wise inference parameters. Chen et al.~\cite{chen2015trainable} further proposed a trainable nonlinear reaction diffusion (TNRD) model through discriminative learning of a compact gradient descent inference step. Recently, Lefkimmiatis~\cite{lefkimmiatis2016non} and Qiao et al.~\cite{Qiao2017} adopted a proximal gradient-based denoising inference from a variational model to incorporate the nonlocal self-similarity prior. {It is worth noting that, apart from MAP inference, Vemulapalli et al.~\cite{Vemulapalli_2016_CVPR} derived an end-to-end trainable patch-based denoising network based on Gaussian Conditional Random Field (GCRF) inference.}

MAP inference guided discriminative learning usually requires much fewer inference steps, and is very efficient in image denoising. It also has clear interpretability because the discriminative architecture is derived from optimization algorithms such as HQS and gradient descent~\cite{schmidt2014shrinkage,barbu2009training,samuel2009learning,sun2011learning,chen2015trainable}. However, the learned priors and inference procedure are limited by the form of MAP model~\cite{tuprints6044}, and generally perform inferior to the state-of-the-art CNN-based denoisers.
For example, the inference of CSF~\cite{schmidt2014shrinkage} is not very flexible since it is strictly derived from the HQS optimization under the field of experts (FoE) framework. The capacity of FoE is however not large enough to fully characterize image priors, which in turn makes CSF less effective.
For these reasons, Kruse et al.~\cite{Kruse_2017_ICCV} generalized CSF for better performance by replacing some modular parts of unrolled inference with more powerful CNN.

\vspace{-0.2cm}
\subsection{Plain Discriminative Learning}

Instead of modeling image priors explicitly, the plain discriminative learning methods learn a direct mapping function to model image prior implicitly. The multi-layer perceptron (MLP) and CNNs have been adopted to learn such priors.
The use of CNN for image denoising can be traced back to~\cite{jain2009natural}, where a five-layer network with sigmoid nonlinearity was proposed. Subsequently, auto-encoder based methods have been suggested for image denoising~\cite{xie2012image,agostinelli2013robust}.
However, early MLP and CNN-based methods are limited in denoising performance and cannot compete with the benchmark BM3D method~\cite{dabov2007image}.

The first discriminative denoising method which achieves comparable performance with BM3D is the plain MLP method proposed by Burger et al.~\cite{burger2012image}.  Benefitted from the advances in deep CNN, Zhang et al.~\cite{zhang2017beyond} proposed a plain denoising CNN (DnCNN) method which achieves state-of-the-art denoising performance. They showed that residual learning and batch normalization~\cite{ioffe2015batch} are particularly useful for the success of denoising.
For a better trade-off between accuracy and speed, Zhang et al.~\cite{zhang2017learning} introduced a 7-layer denoising network with dilated convolution~\cite{yu2015multi} to expand the receptive field of CNN. Mao et al.~\cite{mao2016} proposed a very deep fully convolutional encoding-decoding network with symmetric skip connection for image denoising.
Santhanam et al.~\cite{santhanam2016generalized} developed a recursively branched deconvolutional network (RBDN) for image denoising as well as generic image-to-image regression.
Tai et al.~\cite{tai2017memnet} proposed a very deep persistent memory network (MemNet) by introducing a memory block to mine persistent memory through an adaptive learning process.

Plain discriminative learning has shown better performance than MAP inference guided discriminative learning; however, existing discriminative learning methods have to learn multiple models for handling images with different noise levels, and are incapable to deal with spatially variant noise. To the best of our knowledge, it remains an unaddressed issue to develop a single discriminative denoising model which can handle noise of different levels, even spatially variant noise, in a speed even faster than BM3D.

\section{Proposed Fast and Flexible Discriminative CNN Denoiser}
\label{sec:method}
We present a single discriminative CNN model, namely FFDNet, to achieve the following three objectives:
\begin{itemize}
\item Fast speed: The denoiser is expected to be highly efficient without sacrificing denoising performance.
\item Flexibility: The denoiser is able to handle images with different noise levels and even spatially variant noise.
\item Robustness: The denoiser should introduce no visual artifacts in controlling the trade-off between noise reduction and detail preservation.
\end{itemize}

In this work, we take a tunable noise level map $\emph{\textbf{M}}$ as input to make the denoising model flexible to noise levels. To improve the efficiency of the denoiser, a reversible downsampling operator is introduced to reshape the input image of size $W\times H\times C$ into four downsampled sub-images of size $\frac{W}{2} \times \frac{H}{2} \times 4C$.
Here $C$ is the number of channels, i.e., $C = 1$ for grayscale image and $C = 3$ for color image.
In order to enable the noise level map to robustly control the trade-off between noise
reduction and detail preservation by introducing no visual artifacts, we adopt the orthogonal
initialization method to the convolution filters.

\subsection{Network Architecture}

Fig.~\ref{fig1} illustrates the architecture of FFDNet. The first layer is a reversible downsampling operator which reshapes
a noisy image $\emph{\textbf{y}}$ into four downsampled sub-images.
We further concatenate a tunable noise level map $\emph{\textbf{M}}$ with the downsampled sub-images to form a tensor $\tilde{\emph{\textbf{y}}}$ of size $\frac{W}{2}\times\frac{H}{2}\times (4C + 1)$ as the inputs to CNN.
For spatially invariant AWGN with noise level $\sigma$, $\emph{\textbf{M}}$ is a uniform map with all elements being $\sigma$.

With the tensor $\tilde{\emph{\textbf{y}}}$ as input, the following CNN consists of a series of $3\times3$ convolution layers.
Each layer is composed of three types of operations: Convolution (Conv), Rectified Linear Units (ReLU)~\cite{krizhevsky2012imagenet}, and Batch Normalization (BN)~\cite{ioffe2015batch}.
More specifically, ``Conv+ReLU'' is adopted for the first convolution layer, ``Conv+BN+ReLU'' for the middle layers, and ``Conv'' for the last convolution layer.
Zero-padding is employed to keep the size of feature maps unchanged after each convolution. After the last convolution layer, an upscaling operation is applied as the reverse operator of the downsampling operator applied in the input stage to produce the estimated clean image $\hat{\emph{\textbf{x}}}$ of size $W\times H \times C$.
Different from DnCNN, FFDNet does not predict the noise. The reason is given in Sec.~\ref{section_residual}.
Since FFDNet operates on downsampled sub-images, it is not necessary to employ the dilated convolution~\cite{yu2015multi} to further increase the receptive field.

By considering the balance of complexity and performance, we empirically set the number of convolution layers as 15 for grayscale image and 12 for color image. As to the channels of feature maps, we set 64 for grayscale image and 96 for color image. The reason that we use different settings for grayscale and color images is twofold.
First, since there are high dependencies among the R, G, B channels, using a smaller number of convolution layers encourages the model to exploit the inter-channel dependency.
Second, color image has more channels as input, and hence more feature (i.e., more channels of feature map) is required.
According to our experimental results, increasing the number of feature maps contributes more to the denoising performance on color images. Using different settings for color images, FFDNet can bring an average gain of 0.15dB by PSNR on different noise levels.
As we shall see from Sec.~\ref{section_runtime}, 12-layer FFDNet for color image runs slightly slower than 15-layer FFDNet for grayscale image. Taking both denoising performance and efficiency into account, we set the number of convolution layers as 12 and the number of feature maps as 96 for color image denoising.

\subsection{Noise Level Map}
Let's first revisit the model-based image denoising methods to analyze why they are flexible in handling noises at different levels, which will in turn help us to improve the flexibility of CNN-based denoiser. Most of the model-based denoising methods aim to solve the following problem
\begin{equation}\label{eq3}
  \hat{\emph{\textbf{x}}} = \mathop{\arg\min}\displaystyle_{\emph{\textbf{x}}} ~ \frac{1}{2\sigma^2}\|\emph{\textbf{y}} - \emph{\textbf{x}}\|^2 + \lambda \mathbf{\Phi}(\emph{\textbf{x}}),
\end{equation}
where $\frac{1}{2\sigma^2}\|\emph{\textbf{y}} - \emph{\textbf{x}}\|^2$ is the data fidelity term with noise level $\sigma$,
$\mathbf{\Phi}(\emph{\textbf{x}})$ is the regularization term associated with image prior, and $\lambda$ controls the balance between the data fidelity and regularization terms. It is worth noting that in practice $\lambda$ governs the compromise between noise reduction and detail preservation. When it is too small, much noise will remain; on the opposite, details will be smoothed out along with suppressing noise.

{With some optimization algorithms, the solution of Eqn.~\eqref{eq3} actually defines an implicit function given by
\begin{equation}\label{eq_rf}
\hat{\emph{\textbf{x}}} = \mathcal{F}(\emph{\textbf{y}}, \sigma, \lambda; \Theta).
\end{equation}
Since $\lambda$ can be absorbed into $\sigma$, Eqn.~\eqref{eq_rf} can be rewritten as
\begin{equation}\label{eq_rf2}
\hat{\emph{\textbf{x}}} = \mathcal{F}(\emph{\textbf{y}}, \sigma; \Theta).
\end{equation}}
In this sense, setting noise level $\sigma$ also plays the role of setting $\lambda$ to control the trade-off between noise reduction and detail preservation. In a word, model-based methods are flexible in handling images with various noise levels by simply specifying $\sigma$ in Eqn.~\eqref{eq_rf2}.

According to the above discussion, it is natural to utilize CNN to learn an explicit mapping of Eqn.~\eqref{eq_rf2} which takes the noise image and noise level as input.
However, since the inputs $\emph{\textbf{y}}$ and $\sigma$ have different dimensions, it is not easy to directly feed them into CNN.
Inspired by the patch based denoising methods which actually set $\sigma$ for each patch, we resolve the dimensionality mismatching problem by stretching the noise level $\sigma$ into a noise level map $\emph{\textbf{M}}$.
In $\emph{\textbf{M}}$, all the elements are $\sigma$. As a result, Eqn.~\eqref{eq_rf2} can be further rewritten as
\begin{equation}\label{eq_rf3}
\hat{\emph{\textbf{x}}} = \mathcal{F}(\emph{\textbf{y}}, \emph{\textbf{M}}; \Theta).
\end{equation}
It is worth emphasizing that $\emph{\textbf{M}}$ can be extended to degradation maps with multiple channels for more general noise models such as the multivariate (3D) Gaussian noise model $ \mathcal{N}(\mathbf{0},\mathbf{\Sigma})$ with zero mean and covariance matrix $\mathbf{\Sigma}$ in the RGB color space~\cite{Nam_2016_CVPR}.
As such, a single CNN model is expected to inherit the flexibility of handling noise model with different parameters, even spatially variant noises by noting $\emph{\textbf{M}}$ can be non-uniform.

\subsection{Denoising on Sub-images}

Efficiency is another crucial issue for practical CNN-based denoising. One straightforward idea is to reduce the depth and number of filters. However, such a strategy will sacrifice much the modeling capacity and receptive field of CNN~\cite{zhang2017beyond}.
In~\cite{zhang2017learning}, dilated convolution is introduced to expand receptive field without the increase of network depth, resulting in a 7-layer denoising CNN. Unfortunately, we empirically find that FFDNet with dilated convolution tends to generate artifacts around sharp edges.

Shi et al.~\cite{shi2016real} proposed to extract deep features directly from the low-resolution image for super-resolution, and introduced a sub-pixel convolution layer to improve computational efficiency. In the application of image denoising, we introduce a reversible downsampling layer to reshape the input image into a set of small sub-images. Here the downsampling factor is set to 2 since it can largely improve the speed without reducing modeling capacity. The CNN is deployed on the sub-images, and finally a sub-pixel convolution layer is adopted to reverse the downsampling process.

Denoising on downsampled sub-images can also effectively expand the receptive field which in turn leads to a moderate network depth. For example, the proposed network with a depth of 15 and $3\times3$ convolution will have a large receptive field of $62\times62$.
In contrast, a plain 15-layer CNN only has a receptive field size of 31$\times$31.
We note that the receptive field of most state-of-the-art denoising methods ranges from $35\times35$ to $61\times61$~\cite{zhang2017beyond}.
Further increase of receptive field actually benefits little in improving denoising performance~\cite{levin2011natural}.
What is more, the introduction of subsampling and sub-pixel convolution is effective in reducing the memory burden.

Experiments are conducted to validate the effectiveness of downsampling for balancing denoising accuracy and efficiency on the BSD68 dataset with $\sigma=15$ and $50$. For grayscale image denoising, we train a baseline CNN which has the same depth as FFDNet without downsampling. The comparison of average PSNR values is given as follows: (i) when $\sigma$ is small (i.e., $15$), the baseline CNN slightly outperforms FFDNet by 0.02dB; (ii) when $\sigma$ is large (i.e., $50$), FFDNet performs better than the baseline CNN by 0.09dB. However, FFDNet is nearly 3 times faster and is more memory-friendly than the baseline CNN. As a result, by performing denoising on sub-images, FFDNet significantly improves efficiency while maintaining denoising performance.

\subsection{{Examining the Role of Noise Level Map}}

\begin{figure}[!tbp]
\begin{center}
\subfigure[]
{\includegraphics[width=0.155\textwidth]{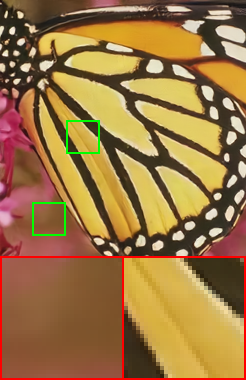}}
\subfigure[]
{\includegraphics[width=0.155\textwidth]{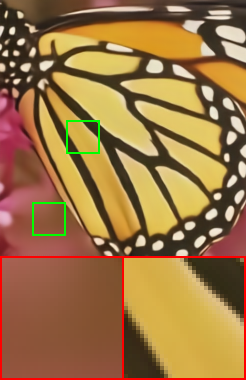}}
\subfigure[]
{\includegraphics[width=0.155\textwidth]{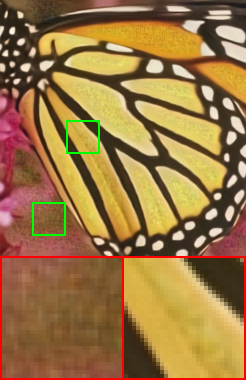}}
\caption{An example to show the importance of guaranteeing the role of noise level map in controlling the trade-off between noise reduction and detail preservation. The input is a noisy image with noise level 25. (a) Result without visual artifacts by matched noise level 25. (b) Result without visual artifacts by mismatched noise level 60. (c) Result with visual artifacts by mismatched noise level 60.}\label{fig_na1}
\end{center}
\end{figure}

By training the model with abundant data units $(\emph{\textbf{y}}, \emph{\textbf{M}}; \emph{\textbf{x}})$, where $\emph{\textbf{M}}$ is exactly the noise level map of $\emph{\textbf{y}}$, the model is expected to perform well when the noise level matches the ground-truth one (see Fig.~\ref{fig_na1}(a)).
On the other hand, in practice, we may need to use the learned model to smooth out some details with a higher noise level map than the ground-truth one (see Fig.~\ref{fig_na1}(b)).
In other words, one may take advantage of the role of $\lambda$ to control the trade-off between noise reduction and detail preservation.
Hence, it is very necessary to further examine whether $\emph{\textbf{M}}$ can play the role of $\lambda$.

Unfortunately, the use of both $\emph{\textbf{M}}$ and $\emph{\textbf{y}}$ as input also increases the difficulty to train the model.
According to our experiments on several learned models, the model may give rise to visual artifacts especially when the input noise level is much higher than the ground-truth one (see Fig.~\ref{fig_na1}(c)), which indicates $\emph{\textbf{M}}$ fails to control the trade-off between noise reduction and detail preservation.
Note that it does not mean all the models suffer from such problem.
One possible solution to avoid this is to regularize the convolution filters. As a widely-used regularization method, orthogonal regularization has proven to be effective in eliminating the correlation between convolution filters, facilitating gradient propagation and improving the compactness of the learned model.
In addition, recent studies have demonstrated the advantage of orthogonal regularization in enhancing the network generalization ability in applications of deep hashing and image classification~\cite{wang2015deep,mhammedi2016efficient,jia2016improving,xie2017all,sun2017svdnet}.
According to our experiments, we empirically find that the orthogonal initialization of the convolution filters~\cite{jia2016improving,2015arXiv151106422M} works well in suppressing the above mentioned visual artifacts.

It is worth emphasising that this section aims to highlight the necessity of guaranteeing the role of $\emph{\textbf{M}}$ in controlling the trade-off between noise reduction and detail preservation rather than proposing a method to avoid the possible visual artifacts caused by noise level mismatch.
In practice, one may retrain the model until $\emph{\textbf{M}}$ plays its role and results in no visual artifacts with a lager noise level.

\subsection{{FFDNet vs. a Single Blind Model}}
So far, we have known that it is possible to learn a single model for blind and non-blind Gaussian denoising, respectively.
And it is of significant importance to clarify their differences.

First, the generalization ability is different. Although the blind model performs favorably for synthetic AWGN removal without knowing the noise level,
it does not generalize well to real noisy images whose noises are much more complex than AWGN (see the results of DnCNN-B in Fig.~\ref{fig_rn1}). Actually, since the CNN model can be
treated as the inference of Eqn.~\eqref{eq3} and the data fidelity term corresponds to the degradation process (or the noise model),
the modeling accuracy of the degradation process is very important for the success of a denoising model.
For example, a model trained for AWGN removal is not expected to be still effective for Poisson noise removal.
By contrast, the non-blind FFDNet model can be viewed as multiple denoisers,
each of which is anchored with a noise level.
Accordingly, it has the ability to control the trade-off between noise removal and detail preservation which in turn facilitates the removal of real noise to some extent (see the results of DnCNN and FFDNet in Fig.~\ref{fig_rn1}).

Second, the performance for AWGN removal is different. The non-blind model with noise level map has moderately better performance for AWGN removal than the blind one (about 0.1dB gain on average for the BSD68 dataset),
possibly because the noise level map provides additional information to the input.
Similar phenomenon has also been recognized in the task of single image super-resolution (SISR)~\cite{riegler2015conditioned}.

Third, the application range is different. In the variable splitting algorithms for general image restoration tasks,
the prior term involves a denoising subproblem with a current noise level~\cite{afonso2010fast,heide2014flexisp,romano2016little}.
Thus, the non-blind model can be easily plugged into variable splitting algorithms to solve various image restoration tasks, such as
image deblurring, SISR, and image inpainting~\cite{chan2016plug,zhang2017learning}. However, the blind model does not have this merit.

\subsection{Residual vs. Non-residual Learning of Plain CNN}\label{section_residual}

It has been pointed out that the integration of residual learning for plain CNN and batch normalization is beneficial to the removal of AWGN
as it eases the training and tends to deliver better performance~\cite{zhang2017beyond}.
The main reason is that the residual (noise) output follows a Gaussian distribution which facilitates the Gaussian normalization step of batch normalization.
The denoising network gains most from such a task-specific merit especially when a single noise level is considered.

In FFDNet, we instead consider a wide range of noise level and introduce a noise level map as input. Thus, it is interesting to revisit the integration of residual learning and batch normalization for plain CNN.
According to our experiments, batch normalization can always accelerate the training of denoising network regardless of the residual or non-residual learning strategy of plain CNN.
In particular, with batch normalization, while the residual learning enjoys a faster convergence than non-residual learning, their final performances after fine-tuning are almost exactly the same.
Some recent works have proposed to train very deep plain networks with nearly the same performance to that with residual learning~\cite{Zagoruyko2017diracnets,xie2017all}.
In fact, when a network is moderately deep (e.g., less than 20), it is feasible to train a plain network without the residual learning strategy by using advanced CNN training and design techniques such as ReLU~\cite{krizhevsky2012imagenet}, batch normalization~\cite{ioffe2015batch} and Adam~\cite{kingma2014adam}.
For simplicity, we do not use residual learning for network design.

\subsection{Un-clipping  vs. Clipping of Noisy Images for Training}\label{section_clip}

In the AWGN denoising literature, there exist two widely-used settings, i.e., un-clipping~\cite{dabov2007image,gu2014weighted,burger2012image,chen2015trainable} and clipping~\cite{schmidt2014shrinkage,Vemulapalli_2016_CVPR}, of synthetic noisy image to evaluate the performance of denoising methods.
The main difference between the two settings lies in whether the noisy image is clipped into the range of 0-255 (or more precisely, quantized into 8-bit format) after adding the noise.

On the one hand, the un-clipping setting which is also the most widely-used setting serves an ideal test bed for evaluating
the denoising methods. This is because most denoising methods assume the noise is ideal AWGN, and
the clipping of noisy input would make the noise characteristics deviate from being AWGN.
Furthermore, in the variable splitting algorithms for solving general image restoration problems, there exists a subproblem which, from a Bayesian perspective, corresponds to a Gaussian denoising problem~\cite{chan2016plug,zhang2017learning}. This further broadens the use of the un-clipping setting.
Thus, unless otherwise specified, FFDNet in this work refers to the model trained on images without clipping or quantization.

On the other hand, since real noisy images are always integer-valued and range-limited,
it has been argued that the clipping setting of noisy image makes the data more realistic~\cite{schmidt2014shrinkage}.
However, when the noise level is high, the noise will be not zero-mean any more due to clipping effects~\cite{Plotz_2017_CVPR}.
This in turn will lead to unreliable denoiser for plugging into the variable splitting algorithms to solve other image restoration problems.

To thoroughly evaluate the proposed method, we also train an FFDNet model with clipping setting of noisy image, namely FFDNet-Clip, for comparison.
During training and testing of FFDNet-Clip, the noisy images are quantized into 8-bit format. Specifically, for a clean image $\textbf{\emph{x}}$, we use the matlab function \texttt{imnoise}$($$\textbf{\emph{x}}$, $\texttt{'gaussian'}$, \texttt{0}, \texttt{$(\frac{\sigma}{255})^2$}$)$ to generate the quantized noisy $\textbf{\emph{y}}$ with noise level $\sigma$.

\section{Experiments}\label{sec:experiments}

\subsection{Dataset Generation and Network Training}\label{section_train}
To train the FFDNet model, we need to prepare a training dataset of input-output pairs $\{(\emph{\textbf{y}}_i, \emph{\textbf{M}}_{i}; \emph{\textbf{x}}_i)\}_{i = 1}^N$. Here, $\emph{\textbf{y}}_i$ is obtained by adding AWGN to latent image $\emph{\textbf{x}}_i$, and $\emph{\textbf{M}}_{i}$ is the noise level map.
The reason to use AWGN to generate the training dataset is two-fold. First, AWGN is a natural choice when there is no specific prior information on noise source. Second, real-world noise can be approximated as locally AWGN~\cite{lee1981refined}. More specifically, FFDNet model is trained on the noisy images $\emph{\textbf{y}}_i = \emph{\textbf{x}}_i + \emph{\textbf{v}}_i$ without quantization to 8-bit integer values.
Though the real noisy images are generally 8-bit quantized, we empirically found that the learned model still works effectively on real noisy images.
For FFDNet-Clip, as mentioned in Sec.~\ref{section_clip}, we use the matlab function \texttt{imnoise} to generate the  quantized noisy image from a clean one.

\begin{table}[!tbp]\footnotesize\arrayrulewidth0.5pt
\caption{Main specifications of the proposed FFDNet}
\center
\begin{tabular}{|p{1.1cm}<{\centering}|p{1.2cm}<{\centering}|p{1.3cm}<{\centering}|p{1.4cm}<{\centering}|p{1.4cm}<{\centering}|}
  \hline\rowcolor[gray]{.9}
  FFDNet models    & Number of layers   & Number of channels   & Noise level range   & Training patch size    \\ \hline
  Grayscale      & 15 &  64   & $[0, 75]$  & 70$\times$70    \\\hline
  Color      & 12  &  96  & $[0, 75]$  & 50$\times$50     \\
  \hline
\end{tabular}
\label{table_network}
\end{table}

We collected a large dataset of source images, including 400 BSD images, 400 images selected from the validation set of ImageNet~\cite{deng2009imagenet}, and the 4,744 images from the Waterloo Exploration Database~\cite{ma2016gmad}.
In each epoch, we randomly crop $N = 128\times 8,000$ patches from these images for training.
The patch size should be larger than the receptive field of FFDNet, and we set it to 70$\times$70 and 50$\times$50 for grayscale images and color images, respectively. The noisy patches are obtained by adding AWGN of noise level $\sigma \in$ $[0, 75]$ to the clean patches.
For each noisy patch, the noise level map is uniform. Since FFDNet is a fully convolutional neural network, it inherits the local connectivity property that the output pixel is determined by the local noisy input and local noise level. Hence, the trained FFDNet naturally has the ability to handle spatially variant noise by specifying a non-uniform noise level map. For clarity, in Table~\ref{table_network} we list the main specifications of the FFDNet models.

\begin{figure*}[!htbp]
\begin{center}
\subfigure[BM3D (26.21dB)]
{\includegraphics[width=0.161\textwidth]{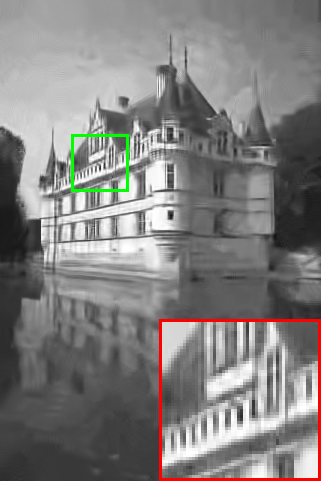}}
\subfigure[WNNM (26.51dB)]
{\includegraphics[width=0.161\textwidth]{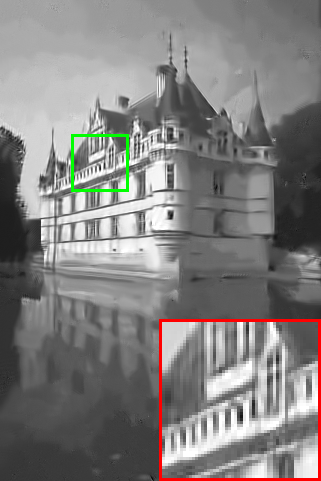}}
\subfigure[MLP (26.54dB)]
{\includegraphics[width=0.161\textwidth]{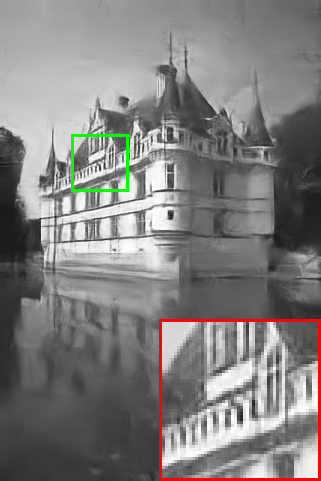}}
\subfigure[TNRD (26.59dB)]
{\includegraphics[width=0.161\textwidth]{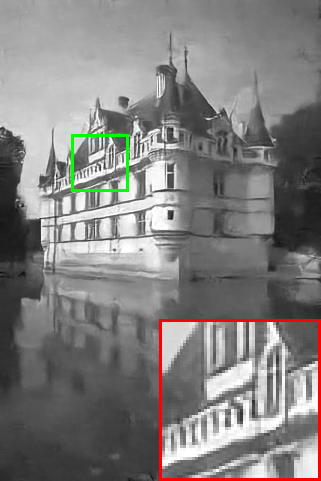}}
\subfigure[DnCNN (26.89dB)]
{\includegraphics[width=0.161\textwidth]{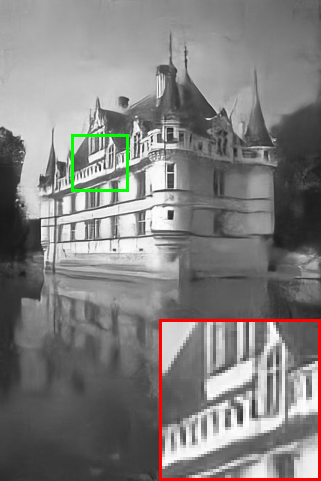}}
\subfigure[FFDNet (26.93dB)]
{\includegraphics[width=0.161\textwidth]{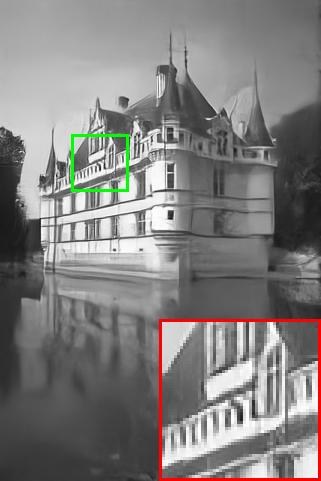}}%
\caption{Denoising results on image ``\emph{102061}'' from the BSD68 dataset with noise level 50 by different methods.}\label{fig3}
\end{center}
\end{figure*}

The ADAM algorithm~\cite{kingma2014adam} is adopted to optimize FFDNet by minimizing the following loss function,
\begin{equation}\label{eq:loss}
  \mathcal{L}(\Theta) = \frac{1}{2N}\sum\nolimits_{i=1}^N\|\mathcal{F}(\emph{\textbf{y}}_i, \emph{\textbf{M}}_{i}; \Theta)  - \emph{\textbf{x}}_i \|^2.
\end{equation}
The learning rate starts from $10^{-3}$ and reduces to $10^{-4}$ when the training error stops decreasing. When the training error keeps unchanged in five sequential epochs, we merge the parameters of each batch normalization into the adjacent convolution filters. Then, a smaller learning rate of $10^{-6}$ is adopted for additional 50 epochs to fine-tune the FFDNet model.
As for the other hyper-parameters of ADAM, we use their default settings. The mini-batch size is set as 128, and the rotation and flip based data augmentation is also adopted during training. The FFDNet models are trained in Matlab (R2015b) environment with MatConvNet package~\cite{vedaldi2015matconvnet} and an Nvidia Titan X Pascal GPU.
The training of a single model can be done in about two days.

To evaluate the proposed FFDNet denoisers on grayscale image denoising, we use BSD68~\cite{roth2005fields} and Set12 datasets to test FFDNet for removing AWGN noise, and use the ``RNI6'' dataset~\cite{lebrun2015noise} to test FFDNet for removing real noise.
The BSD68 dataset consists of 68 images from the separate test set of the BSD300 dataset~\cite{MartinFTM01}. The Set12 dataset is a collection of widely-used testing images. The RNI6 dataset contains 6 real noisy images without ground-truth.
In particular, to evaluate FFDNet-Clip, we use the quantized ``Clip300'' dataset which comprises the 100 images of test set from the BSD300 dataset~\cite{MartinFTM01} and 200 images from PASCALVOC 2012~\cite{Everingham2015} dataset. Note that all the testing images are not included in the training dataset.

As for color image denoising, we employ four datasets, namely CBSD68, Kodak24~\cite{franzen1999kodak}, McMaster~\cite{zhang2011color}, and ``RNI15''~\cite{lebrun2015noise,RNI1}.
The CBSD68 dataset is the corresponding color version of the grayscale BSD68 dataset. The Kodak24 dataset consists of 24 center-cropped images of size 500$\times$500 from the original Kodak dataset. The McMaster dataset is a widely-used dataset for color demosaicing, which contains 18 cropped images of size 500$\times$500.
Compared to the Kodak24 images, the images in McMaster dataset exhibit more saturated colors~\cite{zhang2011color}.
The RNI15 dataset consists of 15 real noisy images. We note that RNI6 and RNI15 cover a variety of real noise types, such as camera noise and JPEG compression noise. Since the ground-truth clean images are unavailable for real noisy images, we thus only provide the visual comparisons on these images. The source codes of FFDNet and its extension to multivariate Gaussian noise are available at \url{https://github.com/cszn/FFDNet}.

\subsection{Experiments on AWGN Removal}

\begin{table*}[!htbp]\footnotesize\arrayrulewidth0.5pt
\caption{The PSNR(dB) results of different  methods on Set12 dataset with noise levels 15, 25 35, 50 and 75. The best two results are highlighted in \textcolor[rgb]{1.00,0.00,0.00}{red} and \textcolor[rgb]{0.00,0.00,1.00}{blue} colors, respectively}
\center
\begin{tabular}{|p{1.55cm}<{\centering}|p{0.8cm}<{\centering}|p{0.8cm}<{\centering}|p{0.8cm}<{\centering}|p{0.8cm}<{\centering}|p{0.8cm}<{\centering}|p{0.8cm}<{\centering}|p{0.8cm}<{\centering}|p{0.8cm}<{\centering}|p{0.8cm}<{\centering}|p{0.8cm}<{\centering}|p{0.8cm}<{\centering}|p{0.8cm}<{\centering}|p{.85cm}<{\centering}|}
  \hline\rowcolor[gray]{.9}
  Images & \scriptsize{\emph{C.man}} & \scriptsize{\emph{House}} & \scriptsize{\emph{Peppers}} & \scriptsize{\emph{Starfish}} & \scriptsize{\emph{Monarch}} &\scriptsize{\emph{Airplane}} & \scriptsize{\emph{Parrot}} & \scriptsize{\emph{Lena}} & \scriptsize{\emph{Barbara}} & \scriptsize{\emph{Boat}} & \scriptsize{\emph{Man}} & \scriptsize{\emph{Couple}} & \scriptsize{\emph{Average}} \\ \hline \hline
    Noise Level & \multicolumn{13}{c|}{$\sigma=15$}   \\ \hline
    BM3D & 31.91 & 34.93 & 32.69 & 31.14 & 31.85 & 31.07 & 31.37 & 34.26 & \textcolor{blue}{33.10} & 32.13 & 31.92 & 32.10 & 32.37  \\\hline
    WNNM& 32.17 & \textcolor{red}{35.13} & 32.99 & 31.82 & \textcolor{blue}{32.71} & 31.39 & 31.62 & 34.27 & \textcolor{red}{33.60} & 32.27 & 32.11 & 32.17 &  32.70  \\\hline
     TNRD& 32.19 & 34.53 & 33.04 & 31.75 & 32.56 & 31.46 & 31.63 & 34.24 & 32.13 & 32.14 & 32.23 & 32.11 & 32.50  \\\hline
     DnCNN& \textcolor{red}{32.61} & 34.97 & \textcolor{red}{33.30} & \textcolor{red}{32.20} & \textcolor{red}{33.09} & \textcolor{red}{31.70} & \textcolor{red}{31.83} & 34.62 & 32.64 & \textcolor{red}{32.42} & \textcolor{red}{32.46} & \textcolor{red}{32.47} &  \textcolor{red}{32.86}  \\\hline
    FFDNet & \textcolor{blue}{32.42} & \textcolor{blue}{35.01} & \textcolor{blue}{33.10} & \textcolor{blue}{32.02} & 32.77 & \textcolor{blue}{31.58} & \textcolor{blue}{31.77} & \textcolor{red}{34.63} & 32.50 & \textcolor{blue}{32.35} & \textcolor{blue}{32.40} & \textcolor{blue}{32.45} &  \textcolor{blue}{32.75}  \\\hline

     Noise Level& \multicolumn{13}{c|}{$\sigma=25$}   \\ \hline
    BM3D & 29.45 & 32.85 & 30.16 & 28.56 & 29.25 & 28.42 & 28.93 & 32.07 & \textcolor{blue}{30.71} & 29.90 & 29.61 & 29.71 &  29.97  \\\hline
     WNNM& 29.64 & \textcolor{blue}{33.22} & 30.42 & 29.03 & 29.84 & 28.69 & 29.15 & 32.24 & \textcolor{red}{31.24} & 30.03 & 29.76 & 29.82 & 30.26   \\\hline
     MLP&  29.61 & 32.56 & 30.30 & 28.82 & 29.61 & 28.82 & 29.25 & 32.25 & 29.54 & 29.97 & 29.88 & 29.73 & 30.03   \\\hline
     TNRD& 29.72 & 32.53 & 30.57 & 29.02 & 29.85 & 28.88 & 29.18 & 32.00 & 29.41 & 29.91 & 29.87 & 29.71 & 30.06    \\\hline
     DnCNN& \textcolor{red}{30.18} & 33.06 & \textcolor{blue}{30.87} & \textcolor{red}{29.41} & \textcolor{red}{30.28} & \textcolor{red}{29.13} & \textcolor{blue}{29.43} & \textcolor{blue}{32.44} & 30.00 & \textcolor{blue}{30.21} & \textcolor{blue}{30.10} & \textcolor{blue}{30.12} &  \textcolor{blue}{30.43} \\\hline
    FFDNet& \textcolor{blue}{30.06} & \textcolor{red}{33.27} & \textcolor{red}{30.79} & \textcolor{blue}{29.33} & \textcolor{blue}{30.14} & \textcolor{blue}{29.05} & \textcolor{red}{29.43} & \textcolor{red}{32.59} & 29.98 & \textcolor{red}{30.23} & \textcolor{red}{30.10} & \textcolor{red}{30.18} &  \textcolor{red}{30.43}  \\\hline

 Noise Level& \multicolumn{13}{c|}{$\sigma=35$}   \\ \hline
    BM3D & 27.92 & 31.36 & 28.51 & 26.86 & 27.58 & 26.83 & 27.40 & 30.56 & \textcolor{blue}{28.98} & 28.43 & 28.22 & 28.15 &  28.40  \\\hline
     WNNM& 28.08 & \textcolor{blue}{31.92} & 28.75 & 27.27 & 28.13 & 27.10 & 27.69 & 30.73 & \textcolor{red}{29.48} & 28.54 & 28.33 & 28.24 & 28.69   \\\hline
     MLP& 28.08 & 31.18 & 28.54 & 27.12 & 27.97 & 27.22 & 27.72 & 30.82 & 27.62 & 28.53 & 28.47 & 28.24 & 28.46   \\\hline
      DnCNN& \textcolor{red}{28.61} & 31.61 & \textcolor{blue}{29.14} & \textcolor{blue}{27.53} & \textcolor{blue}{28.51} & \textcolor{blue}{27.52} & \textcolor{blue}{27.94} & \textcolor{blue}{30.91} & 28.09 & \textcolor{blue}{28.72} & \textcolor{blue}{28.66} & \textcolor{blue}{28.52} &  \textcolor{blue}{28.82}  \\\hline
    FFDNet& \textcolor{blue}{28.54} & \textcolor{red}{31.99} & \textcolor{red}{29.18} & \textcolor{red}{27.58} & \textcolor{red}{28.54} & \textcolor{red}{27.47} & \textcolor{red}{28.02} & \textcolor{red}{31.20} & 28.29 & \textcolor{red}{28.82} & \textcolor{red}{28.70} & \textcolor{red}{28.68} &  \textcolor{red}{28.92}  \\\hline

     Noise Level& \multicolumn{13}{c|}{$\sigma=50$}   \\ \hline
    BM3D & 26.13 & 29.69 & 26.68 & 25.04 & 25.82 & 25.10 & 25.90 & 29.05 & \textcolor{blue}{27.22} & 26.78 & 26.81 & 26.46 &  26.72  \\\hline
    WNNM& 26.45 & \textcolor{blue}{30.33} & 26.95 & 25.44 & 26.32 & 25.42 & 26.14 & 29.25 & \textcolor{red}{27.79} & 26.97 & 26.94 & 26.64 & 27.05   \\\hline
     MLP& 26.37 & 29.64 & 26.68 & 25.43 & 26.26 & 25.56 & 26.12 & 29.32 & 25.24 & 27.03 & 27.06 & 26.67 &  26.78  \\\hline
     TNRD& 26.62 & 29.48 & 27.10 & 25.42 & 26.31 & 25.59 & 26.16 & 28.93 & 25.70 & 26.94 & 26.98 & 26.50 & 26.81   \\\hline
     DnCNN& \textcolor{blue}{27.03} & 30.00 & \textcolor{blue}{27.32} & \textcolor{blue}{25.70} & \textcolor{blue}{26.78} & \textcolor{blue}{25.87} & \textcolor{blue}{26.48} & \textcolor{blue}{29.39} & 26.22 & \textcolor{blue}{27.20} & \textcolor{blue}{27.24} & \textcolor{blue}{26.90} &  \textcolor{blue}{27.18}  \\\hline
     FFDNet& \textcolor{red}{27.03} & \textcolor{red}{30.43} & \textcolor{red}{27.43} & \textcolor{red}{25.77} & \textcolor{red}{26.88} & \textcolor{red}{25.90} & \textcolor{red}{26.58} & \textcolor{red}{29.68} & 26.48 & \textcolor{red}{27.32} & \textcolor{red}{27.30} & \textcolor{red}{27.07} &  \textcolor{red}{27.32}  \\\hline

    Noise Level& \multicolumn{13}{c|}{$\sigma=75$}   \\ \hline
    BM3D & 24.32 & 27.51 & 24.73 & 23.27 & 23.91 & 23.48 & 24.18 & 27.25 & \textcolor{blue}{25.12} & 25.12 & 25.32 & 24.70 &  24.91  \\\hline
    WNNM& 24.60 & \textcolor{blue}{28.24} & 24.96 & 23.49 & 24.31 & 23.74 & 24.43 & 27.54 & \textcolor{red}{25.81} & 25.29 & 25.42 & 24.86 &  \textcolor{blue}{25.23}  \\\hline
     MLP& 24.63 & 27.78 & 24.88 & 23.57 & 24.40 & 23.87 & 24.55 & \textcolor{blue}{27.68} & 23.39 & 25.44 & 25.59 & \textcolor{blue}{25.02} &  25.07  \\\hline
      DnCNN& \textcolor{blue}{25.07} & 27.85 & \textcolor{blue}{25.17} & \textcolor{blue}{23.64} & \textcolor{blue}{24.71} & \textcolor{blue}{24.03} & \textcolor{blue}{24.71} & 27.54 & 23.63 & \textcolor{blue}{25.47} & \textcolor{blue}{25.64} & 24.97 &  25.20  \\\hline
     FFDNet& \textcolor{red}{25.29} & \textcolor{red}{28.43} & \textcolor{red}{25.39} & \textcolor{red}{23.82} & \textcolor{red}{24.99} & \textcolor{red}{24.18} & \textcolor{red}{24.94} & \textcolor{red}{27.97} & 24.24 & \textcolor{red}{25.64} & \textcolor{red}{25.75} & \textcolor{red}{25.29} &  \textcolor{red}{25.49}  \\\hline
\end{tabular}
\label{table_set12}
\end{table*}

\begin{table}[!bp]\footnotesize\arrayrulewidth0.5pt
\caption{The average PSNR(dB) results of different methods on BSD68 with noise levels 15, 25 35, 50 and 75}
\center
\begin{tabular}{|p{1.2cm}<{\centering}|p{.75cm}<{\centering}|p{.75cm}<{\centering}|p{.75cm}<{\centering}|p{.75cm}<{\centering}|p{.75cm}<{\centering}|p{.75cm}<{\centering}|p{.75cm}<{\centering}|}
  \hline\rowcolor[gray]{.9}
  Methods &  BM3D&  WNNM& MLP & TNRD   &  DnCNN &   FFDNet \\ \hline
  $\sigma = 15$ & 31.07 &  31.37& -- &31.42  &  31.72 & 31.63  \\\hline
  $\sigma = 25$ & 28.57 &  28.83  & 28.96&28.92 & 29.23   & 29.19\\\hline
  $\sigma = 35$ & 27.08 &  27.30  & 27.50 &-- & 27.69   & 27.73\\\hline
  $\sigma = 50$ & 25.62 &  25.87  & 26.03 &25.97 &  26.23 & 26.29 \\\hline
  $\sigma = 75$ & 24.21 &  24.40  & 24.59 &-- & 24.64   & 24.79\\
  \hline
\end{tabular}
\label{table_bsd68}
\end{table}

\begin{table}[!bp]\footnotesize\arrayrulewidth0.5pt
\caption{The average PSNR(dB) results of different methods on Clip300 with noise levels 15, 25 35, 50 and 60}
\center
\begin{tabular}{|p{1.75cm}<{\centering}|p{.82cm}<{\centering}|p{.82cm}<{\centering}|p{.82cm}<{\centering}|p{.82cm}<{\centering}|p{.82cm}<{\centering}|p{.82cm}<{\centering}|}
  \hline\rowcolor[gray]{.9}
  Methods    & $\sigma$ = 15    & $\sigma$ = 25   & $\sigma$ = 35      & $\sigma$ = 50     & $\sigma$ = 60    \\ \hline
  DCGRF      & 31.35 &  28.67   & 27.08  & 25.38  &  24.45  \\\hline
  RBDN     & 31.05  &  28.77  & 27.31  & 25.80  & 23.25   \\\hline
  FFDNet-Clip& 31.68   &  29.25 & 27.75  &26.25 & 25.51   \\
  \hline
\end{tabular}
\label{table_set300}
\end{table}

\begin{table}[!bp]\footnotesize\arrayrulewidth0.5pt
\caption{The average PSNR(dB) results of CBM3D, CDnCNN and FFDNet on CBSD68, Kodak24 and McMaster datasets with noise levels 15, 25 35, 50 and 75}
\center
\begin{tabular}{|p{1cm}<{\centering}|p{0.96cm}<{\centering}|p{.76cm}<{\centering}|p{.76cm}<{\centering}|p{.76cm}<{\centering}|p{.76cm}<{\centering}|p{.76cm}<{\centering}|p{.76cm}<{\centering}|}
  \hline\rowcolor[gray]{.9}
  Datasets & Methods &$\sigma$$=$15&  $\sigma$$=$25& $\sigma$$=$35  & $\sigma$$=$50 & $\sigma$$=$75 \\ \hline
   &CBM3D    & 33.52    & 30.71 &  28.89 & 27.38 & 25.74 \\
 CBSD68  &{CDnCNN} &33.89 &  31.23    & 29.58 & 27.92   & 24.47\\
  &FFDNet &33.87 &  31.21    & 29.58 & 27.96   & 26.24\\\hline

    & CBM3D   & 34.28    & 31.68 &  29.90 & 28.46 & 26.82 \\
    Kodak24 &{CDnCNN} &34.48 &  32.03    & 30.46 & 28.85   & 25.04\\
   & FFDNet&34.63 &  32.13    & 30.57 & 28.98   & 27.27\\\hline

  & CBM3D   & 34.06    & 31.66 &  29.92 & 28.51 & 26.79 \\
McMaster &{CDnCNN} &33.44 &  31.51    & 30.14 & 28.61  & 25.10\\
  & FFDNet  &34.66 &  32.35   & 30.81 & 29.18   & 27.33\\\hline
\end{tabular}
\label{table3}
\end{table}

In this subsection, we test FFDNet on noisy images corrupted by spatially invariant AWGN. For grayscale image denoising, we mainly compare FFDNet with state-of-the-art methods BM3D~\cite{dabov2007image}, WNNM~\cite{gu2014weighted}, MLP~\cite{burger2012image}, TNRD~\cite{chen2015trainable}, and DnCNN~\cite{zhang2017beyond}.
Note that BM3D and WNNM are two representative model-based methods based on nonlocal self-similarity prior, whereas TNRD, MLP and DnCNN are discriminative learning based methods. Tables~\ref{table_set12} and~\ref{table_bsd68} report the PSNR results on BSD68 and Set12 datasets, respectively.
We also use two CNN-based denoising methods, i.e., RED30~\cite{mao2016} and MemNet~\cite{tai2017memnet}, for further comparison. Their PSNR results on BSD68 dataset with noise level 50 are 26.34dB and 26.35dB, respectively. Note that RED30 and MemNet are trained on a specific noise level and are less efficient than DnCNN.
From Tables~\ref{table_set12} and~\ref{table_bsd68}, one can have the following observations.

First, FFDNet surpasses BM3D by a large margin and outperforms WNNM, MLP and TNRD by about 0.2dB for a wide range of noise levels on BSD68. Second, FFDNet is slightly inferior to DnCNN when the noise level is low (e.g., $\sigma \leq 25$), but gradually outperforms DnCNN with the increase of noise level (e.g., $\sigma > 25$).
This phenomenon may be resulted from the trade-off between receptive field size and modeling capacity. FFDNet has a larger receptive field than DnCNN, thus favoring for removing strong noise, while DnCNN has better modeling capacity which is beneficial for denoising images with lower noise level. Third, FFDNet outperforms WNNM on images such as ``\emph{House}'', while it is inferior to WNNM on image ``\emph{Barbara}''.
This is because ``\emph{Barbara}'' has a rich amount of repetitive structures, which can be effectively exploited by nonlocal self-similarity based WNNM method. The visual comparisons of different methods are given in Fig.~\ref{fig3}.
Overall, FFDNet produces the best perceptual quality of denoised images.

To evaluate FFDNet-Clip, Table~\ref{table_set300} shows the PSNR comparison with DCGRF~\cite{Vemulapalli_2016_CVPR} and RBDN~\cite{santhanam2016generalized} on the Clip300 dataset.
It can be seen that FFDNet-Clip with matched noise level achieves better performance than DCGRF and RBDN, showing that FFDNet performs well under the clipping setting.
We also tested FFDNet-Clip on BSD68 dataset with clipping setting, it has been found that the PSNR result is similar to that of FFDNet with un-clipping setting.

\begin{figure}[!tb]
\begin{center}
\subfigure[CBM3D (25.49dB)]
{\includegraphics[height=0.24\textwidth]{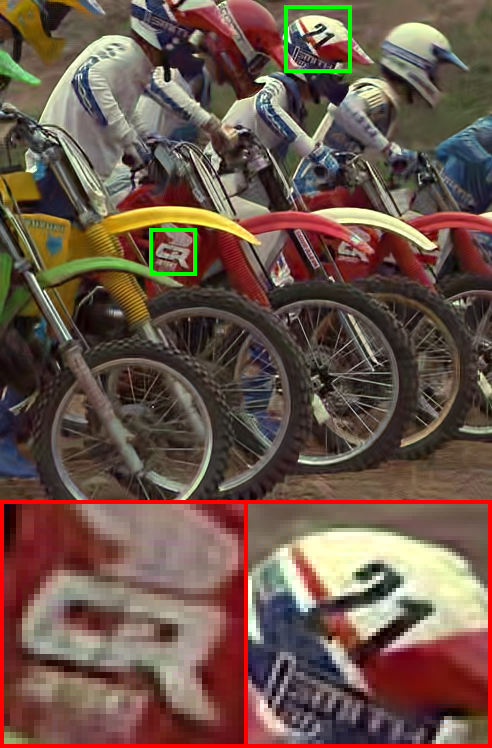}}
\subfigure[CDnCNN (26.19dB)]
{\includegraphics[height=0.24\textwidth]{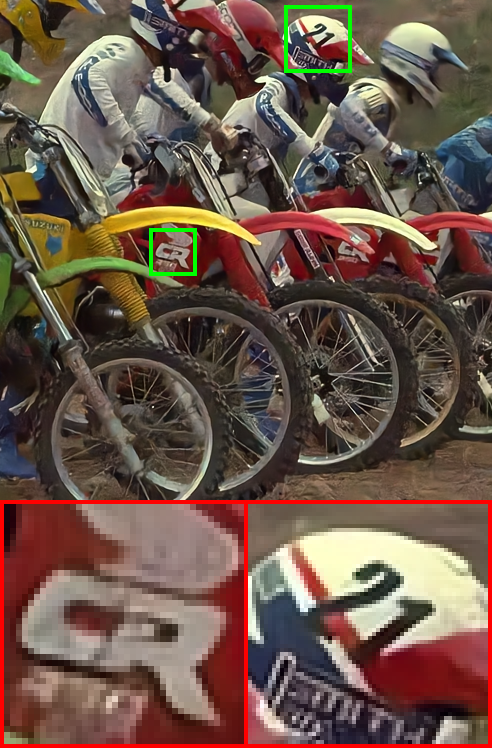}}
\subfigure[FFDNet (26.28dB)]
{\includegraphics[height=0.24\textwidth]{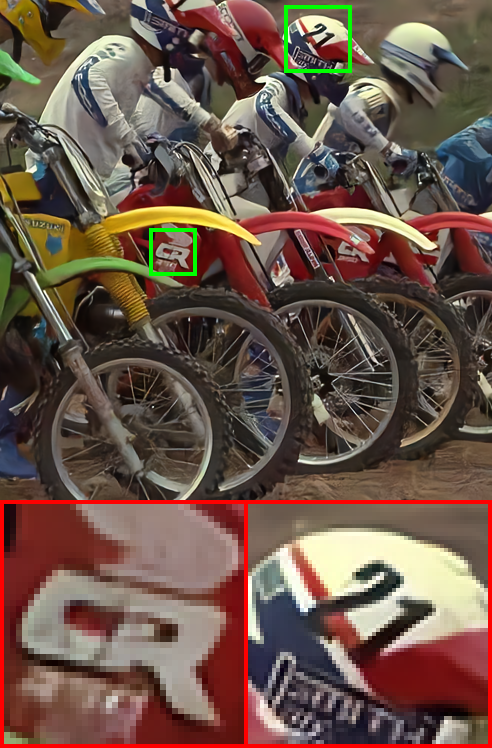}}

\caption{Color image denoising results by CBM3D, CDnCNN and FFDNet on noise level $\sigma = \text{50}$.}\label{fig5}
\end{center}
\end{figure}

For color image denoising, we compare FFDNet with CBM3D~\cite{dabov2007image} and CDnCNN~\cite{zhang2017beyond}.
Table~\ref{table3} reports the performance of different methods on CBSD68, Kodak24, and McMaster datasets, and
Fig.~\ref{fig5} presents the visual comparisons. It can be seen that FFDNet consistently outperforms CBM3D on different noise levels in terms of both quantitative and qualitative evaluation, and has competing performance with CDnCNN.

\subsection{Experiments on Spatially Variant AWGN Removal}
We then test the flexibility of FFDNet to deal with spatially variant AWGN. To synthesize spatially variant AWGN, we first generate an AWGN image $\emph{\textbf{v}}_1$ with unit standard deviation and a noise level map $\emph{\textbf{M}}$ of the same size. Then, element-wise multiplication is applied on $\emph{\textbf{v}}_1$ and $\emph{\textbf{M}}$ to produce the spatially variant AWGN, i.e., $\emph{\textbf{v}} = \emph{\textbf{v}}_1 \odot \emph{\textbf{M}}$.
In the denoising stage, we take the bilinearly downsampled noise level map as the input to FFDNet. Since the noise level map is spatially smooth, the use of downsampled noise level map generally has very little effect on the final denoising performance.

\begin{figure}[!htbp]
\setlength{\abovecaptionskip}{0pt}
\setlength{\belowcaptionskip}{0pt}
\centering
\subfigure[]{
\begin{minipage}[c]{0.15\textwidth}
\centering
  \includegraphics[width=.8\textwidth]{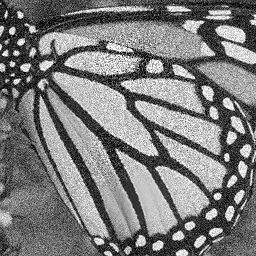}

\end{minipage}%
}
\subfigure[]{
\begin{minipage}[c]{0.15\textwidth}
\centering
  \includegraphics[width=.99\textwidth]{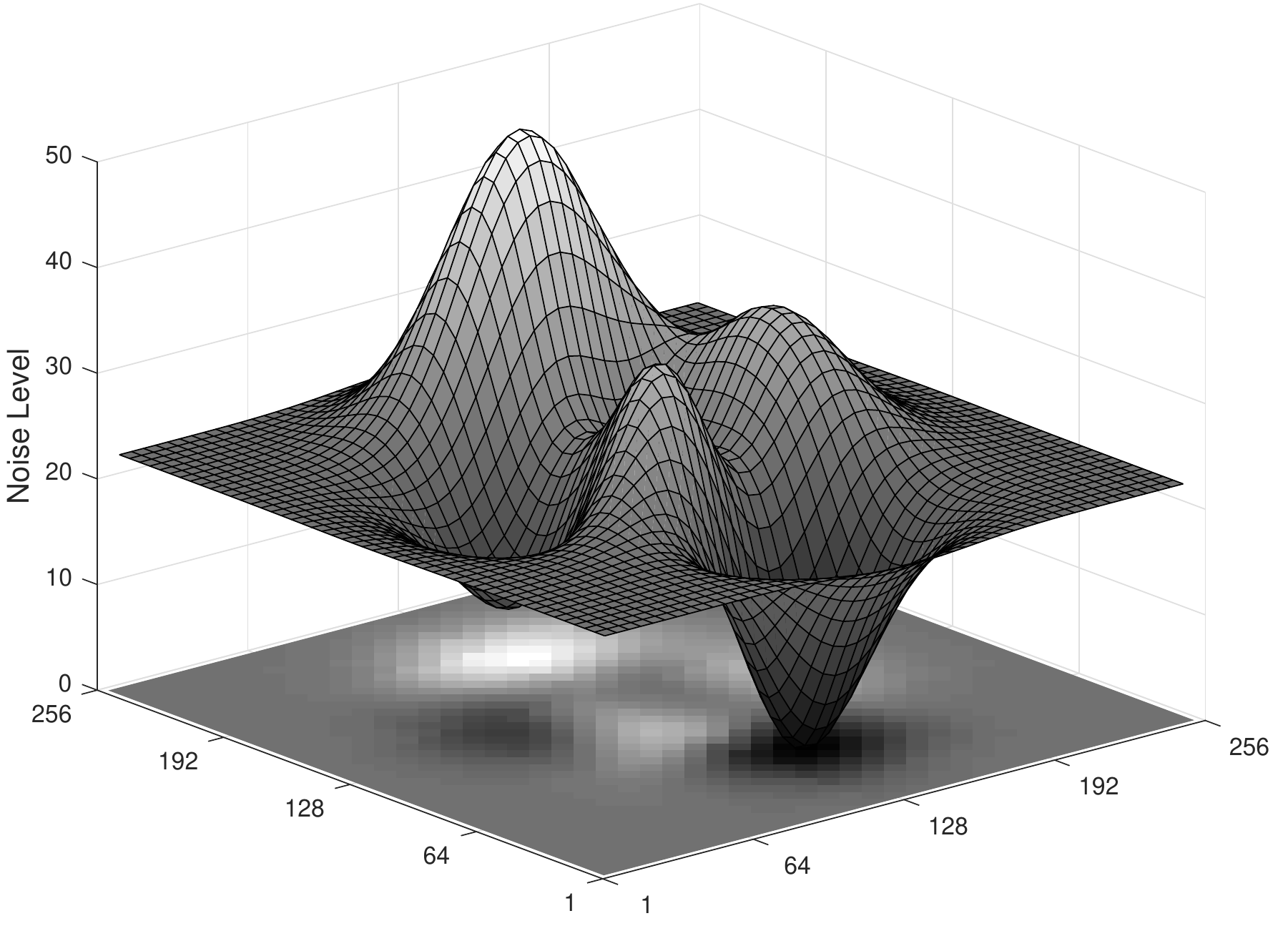}

  \includegraphics[width=.8\textwidth]{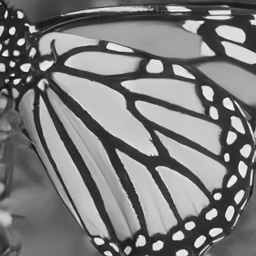}

\end{minipage}%
}
\subfigure[]{
\begin{minipage}[c]{0.15\textwidth}
\centering

  \includegraphics[width=.99\textwidth]{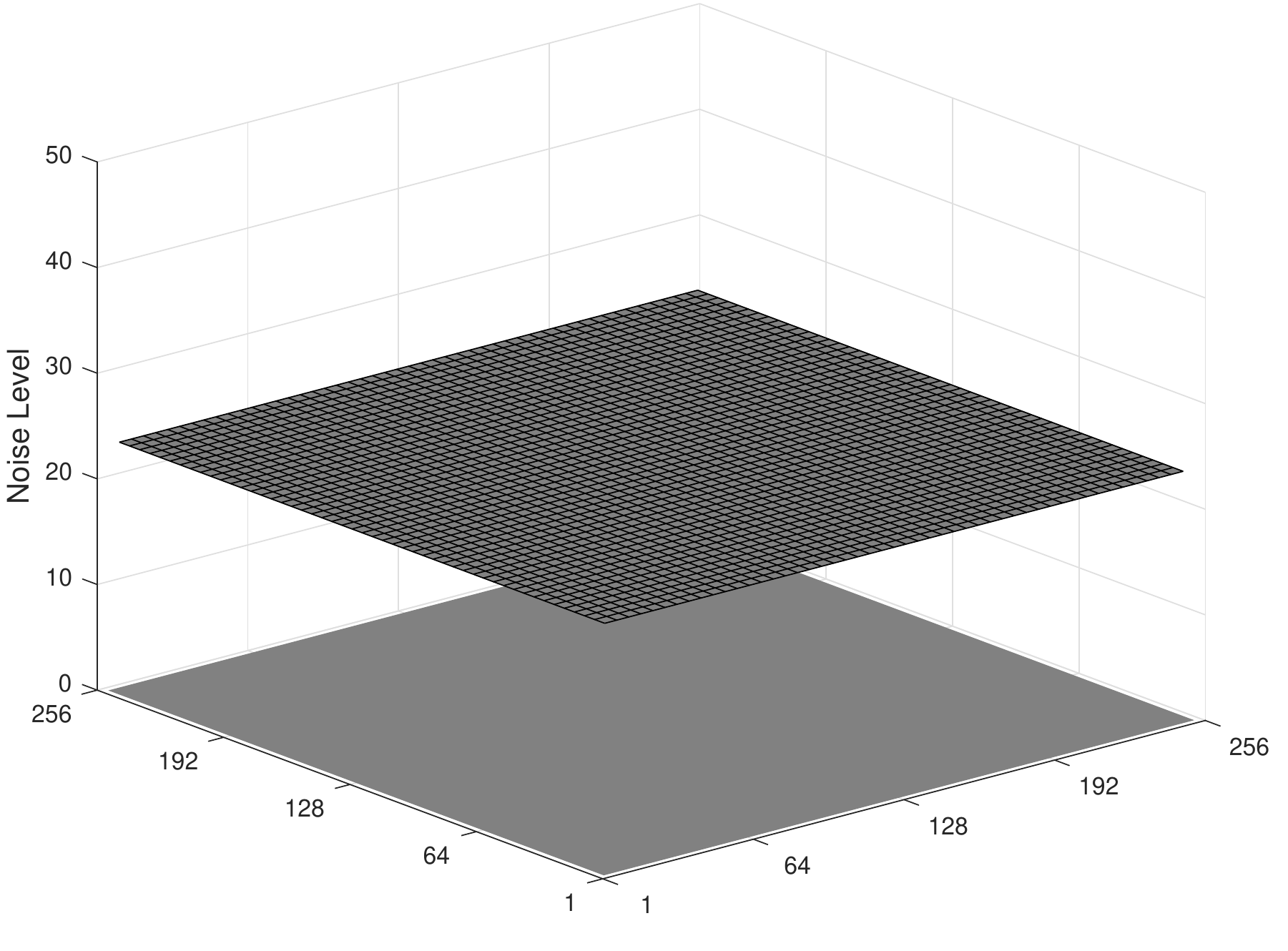}

  \includegraphics[width=.8\textwidth]{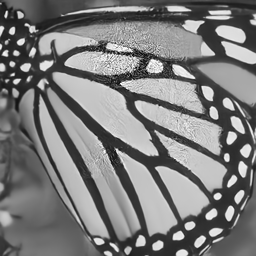}
\end{minipage}%
}
\caption{Examples of FFDNet on removing spatially variant AWGN. (a) Noisy image (20.55dB) with spatially variant AWGN. (b) Ground-truth noise level map and corresponding denoised image (30.08dB) by FFDNet; (c) Uniform noise level map constructed by using the mean value of ground-truth noise level map and corresponding denoised image (27.45dB) by FFDNet.}
\label{fig_sv1}
\end{figure}

Fig.~\ref{fig_sv1} gives an example to show the effectiveness of FFDNet on removing spatially variant AWGN. We do not compare FFDNet with other methods because no state-of-the-art AWGN denoising method can be readily extended to handle spatially variant AWGN. From Fig.~\ref{fig_sv1}, one can see that FFDNet with non-uniform noise level map is flexible and powerful to remove spatially variant AWGN.
In contrast, FFDNet with uniform noise level map would fail to remove strong noise at the region with higher noise level while smoothing out the details at the region with lower noise level.

\subsection{Experiments on Noise Level Sensitivity}

In practical applications, the noise level map may not be accurately estimated from the noisy observation, and mismatch between the input and real noise levels is inevitable. If the input noise level is lower than the real noise level, the noise cannot be completely removed. Therefore, users often prefer to set a higher noise level to remove more noise. However, this may also remove some image details together with noise. A practical denoiser should tolerate certain mismatch of noise levels. In this subsection, we evaluate FFDNet in comparison with benchmark BM3D and DnCNN by varying different input noise levels for a given ground-truth noise level.

\begin{figure}[!bp]
\begin{center}
\subfigure
{\includegraphics[width=0.49\textwidth]{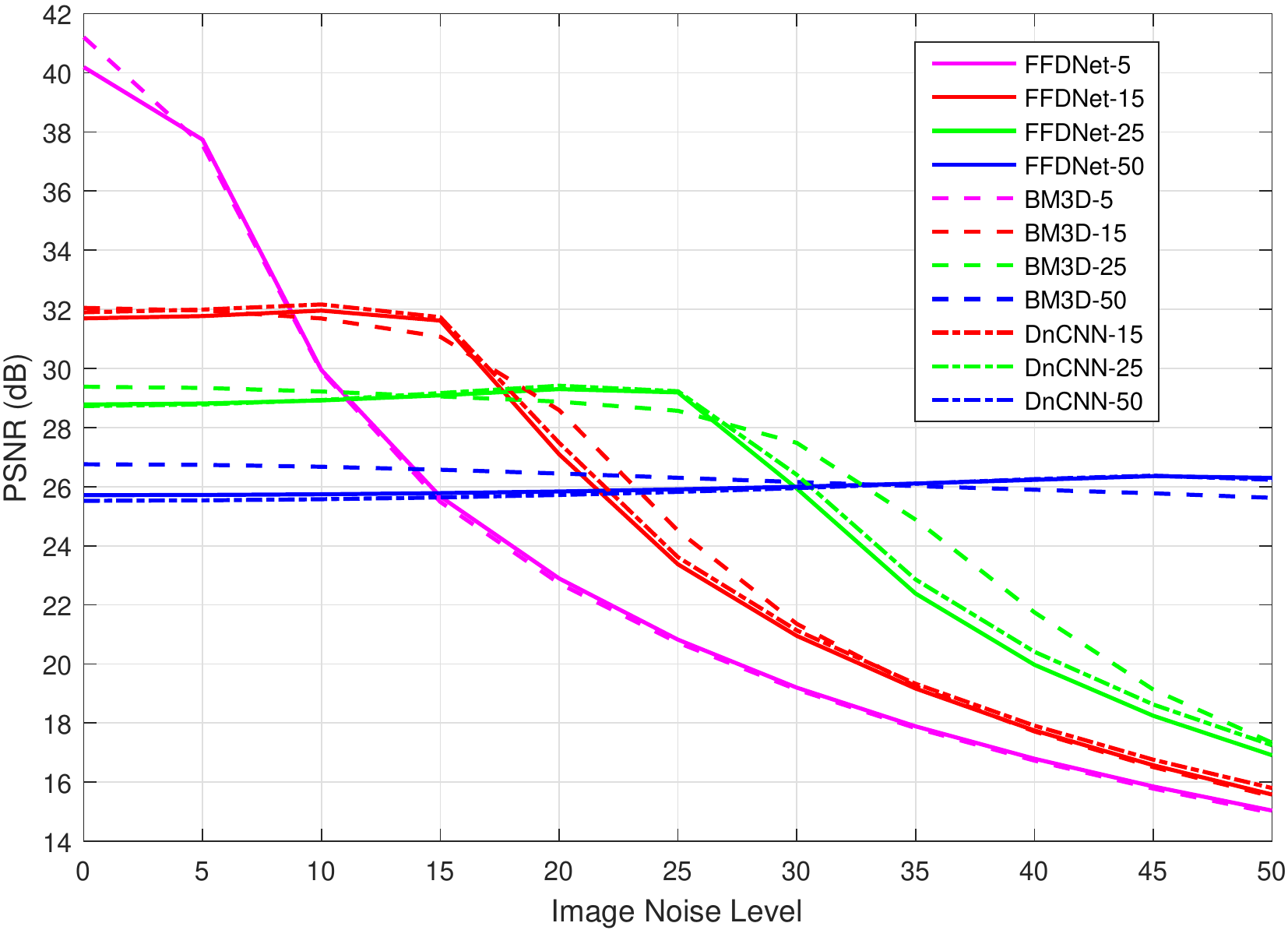}}
\caption{Noise level sensitivity curves of BM3D, DnCNN and FFDNet. The averaged PSNR results are evaluated on BSD68.}\label{fig_nm1}
\end{center}
\end{figure}

\begin{figure*}[!htbp]
\begin{center}
\subfigure[]
{\includegraphics[width=0.16\textwidth]{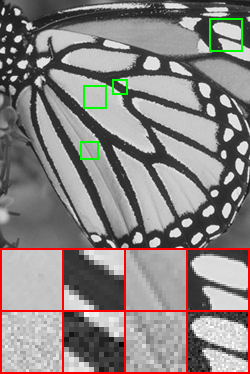}}
\subfigure[]
{\includegraphics[width=0.16\textwidth]{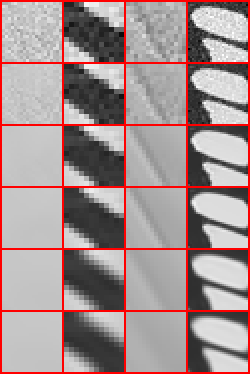}}
\subfigure[]
{\includegraphics[width=0.16\textwidth]{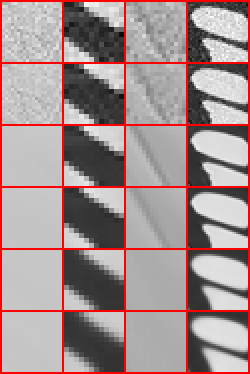}}
\subfigure[]
{\includegraphics[width=0.16\textwidth]{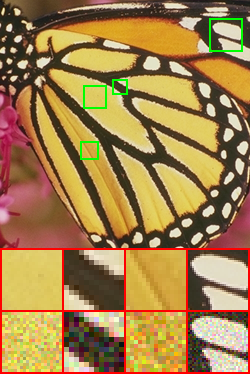}}
\subfigure[]
{\includegraphics[width=0.16\textwidth]{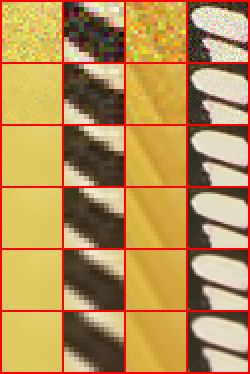}}
\subfigure[]
{\includegraphics[width=0.16\textwidth]{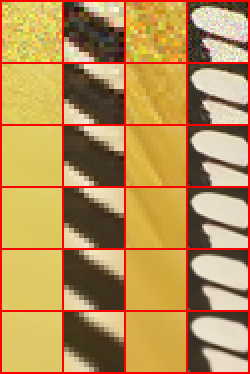}}
\caption{Visual comparisons between FFDNet and BM3D/CBM3D by setting different input noise levels to denoise a noisy image. (a) From top to bottom: ground-truth image, four clean zoom-in regions, and the corresponding noisy regions (AWGN, noise level 15). (b) From top to bottom: denoising results by BM3D with input noise levels 5, 10, 15, 20, 50, and 75, respectively. (c) Results by FFDNet with the same settings as in (b). (d) From top to bottom: ground-truth image, four clean zoom-in regions, and the corresponding noisy regions (AWGN, noise level 25). (e) From top to bottom: denoising results by CBM3D with input noise levels 10, 20, 25, 30, 45 and 60, respectively. (f) Results by FFDNet with the same settings as in (e).}\label{fig_nm2}
\end{center}\vspace{-0.4cm}
\end{figure*}

Fig.~\ref{fig_nm1} illustrates the noise level sensitivity curves of BM3D, DnCNN and FFDNet. Different methods with different input noise levels (e.g., ``FFDNet-15'' represents FFDNet with input noise level fixed as 15) are evaluated on BSD68 images with noise level ranging from 0 to 50.
Fig.~\ref{fig_nm2} shows the visual comparisons between BM3D/CBM3D and FFDNet by setting different input noise levels to denoise a noisy image. Four typical image structures, including flat region, sharp edge, line with high contrast, and line with low contrast, are selected for visual comparison to investigate the noise level sensitivity of BM3D and FFDNet.
From Figs.~\ref{fig_nm1} and~\ref{fig_nm2}, we have the following observations.
\begin{itemize}
   \item On all noise levels, FFDNet achieves similar denoising results to BM3D and DnCNN when their input noise levels are the same.

   \item With the fixed input noise level, for all the three methods, the PSNR value tends to stay the same when the ground-truth noise level is lower, and begins to decrease when the ground-truth noise level is higher.

   \item The best visual quality is obtained when the input noise level matches the ground-truth one. BM3D and FFDNet produce similar visual results with lower input noise levels, while they exhibit certain difference with higher input noise levels. Both of them will smooth out noise in flat regions, and gradually smooth out image structures with the increase of input noise levels. Particularly, FFDNet may wipe out some low contrast line structure, whereas BM3D can still preserve the mean patch regardless of the input noise levels due to its use of nonlocal information.

   \item Using a higher input noise level can generally produce better visual results than using a lower one. In addition, there is no much visual difference when the input noise level is a little higher than the ground-truth one.
\end{itemize}

According to above observations, FFDNet exhibits similar noise level sensitivity performance to BM3D and DnCNN in balancing noise reduction and detail preservation. When the ground-truth noise level is unknown, it is more preferable to set a larger input noise level than a lower one to remove noise with better perceptual quality.

\subsection{Experiments on Real Noisy Images}

In this subsection, real noisy images are used to further assess the practicability of FFDNet. However, such an evaluation is difficult to conduct due to the following reasons. (i) Both the ground-truth clean image and noise level are unknown for real noisy image. (ii) The real noise comes from various sources such as camera imaging pipeline (e.g., shot noise, amplifier noise and quantization noise), scanning, lossy compression and image resizing~\cite{liu2008automatic,colom2014non}, and it is generally non-Gaussian, spatially variant, and signal-dependent. As a result, the AWGN assumption in many denoising algorithms does not hold, and the associated noise level estimation methods do not work well for real noisy images.

Instead of adopting any noise level estimation methods, we adopt an interactive strategy to handle real noisy images. First of all, we empirically found that the assumption of spatially invariant noise usually works well for most real noisy images. We then employ a set of typical input noise levels to produce multiple outputs, and select the one which has best trade-off between noise reduction and detail preservation.
Second, the spatially variant noise in most real-world images is signal-dependent. In this case, we first sample several typical regions of distinct colors. For each typical region, we apply different noise levels with an interval of 5, and choose the best noise level by observing the denoising results. The noise levels at other regions are then interpolated from the noise levels of the typical regions to constitute an approximated non-uniform noise level map. Our FFDNet focuses on non-blind denoising and assumes the noise level map is known. In practice, some advanced noise level estimation methods~\cite{liu2008automatic,azzari2014} can be adopted to assist the estimation of noise level map.
In our following experiments, unless otherwise specified, we assume spatially invariant noise for the real noisy images.

\begin{figure*}[!htbp]\setcounter{subfigure}{-30}
\begin{center}
\subfigure
{\includegraphics[height=0.181\textwidth]{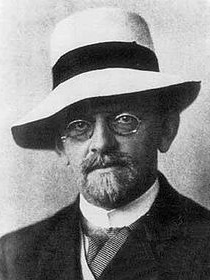}}
\subfigure
{\includegraphics[height=0.181\textwidth]{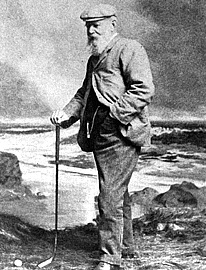}}
\subfigure
{\includegraphics[height=0.181\textwidth]{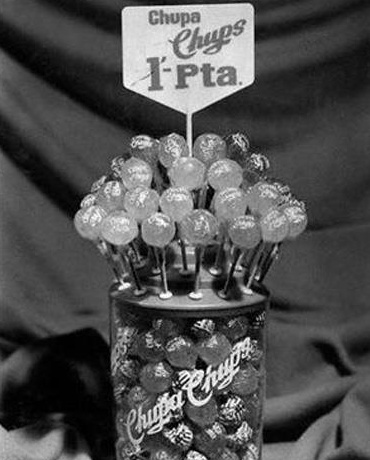}}
\subfigure
{\includegraphics[height=0.181\textwidth]{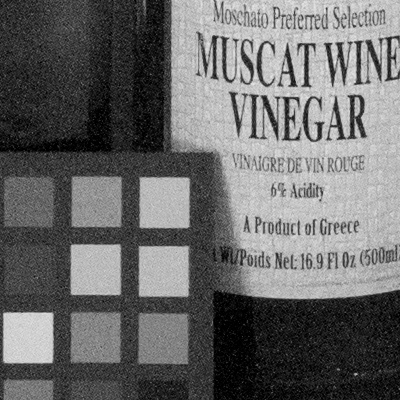}}
\subfigure
{\includegraphics[height=0.181\textwidth]{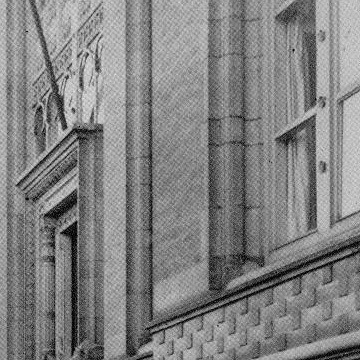}}
\subfigure
{\includegraphics[height=0.181\textwidth]{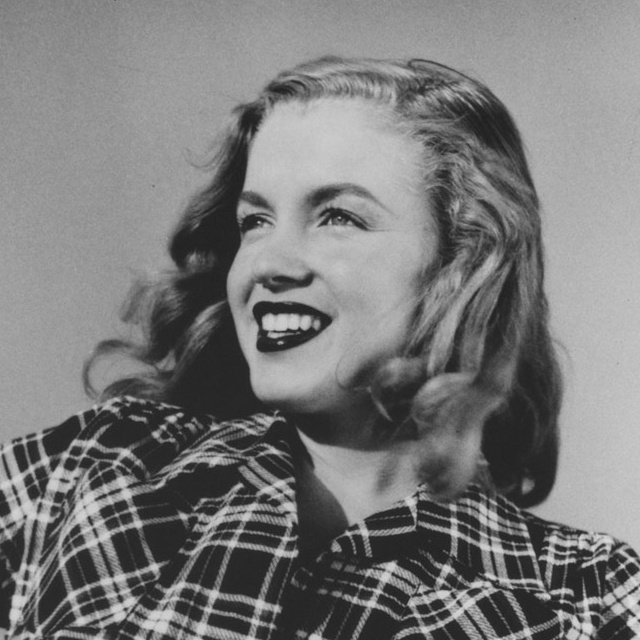}}

\subfigure
{\includegraphics[height=0.181\textwidth]{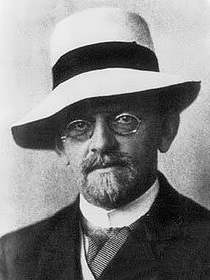}}
\subfigure
{\includegraphics[height=0.181\textwidth]{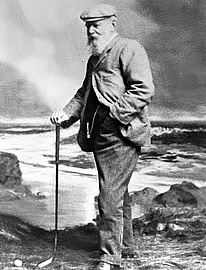}}
\subfigure
{\includegraphics[height=0.181\textwidth]{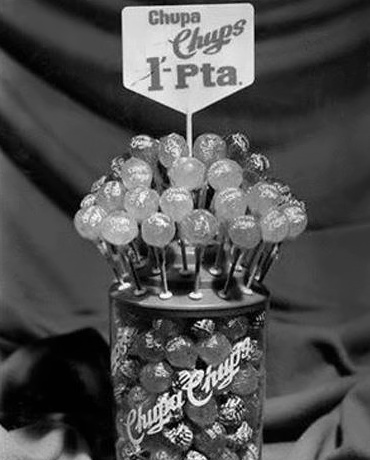}}
\subfigure
{\includegraphics[height=0.181\textwidth]{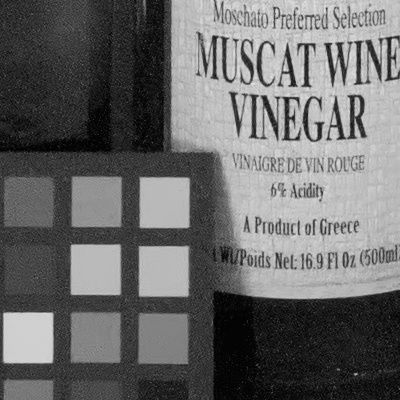}}
\subfigure
{\includegraphics[height=0.181\textwidth]{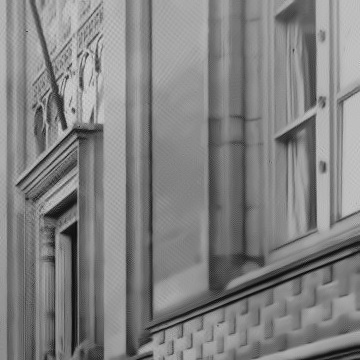}}
\subfigure
{\includegraphics[height=0.181\textwidth]{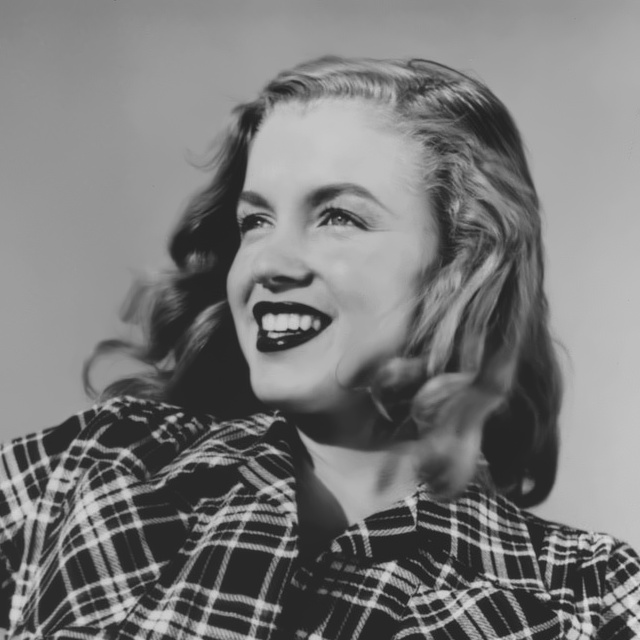}}

\subfigure
{\includegraphics[height=0.181\textwidth]{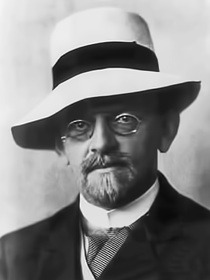}}
\subfigure
{\includegraphics[height=0.181\textwidth]{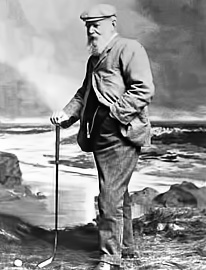}}
\subfigure
{\includegraphics[height=0.181\textwidth]{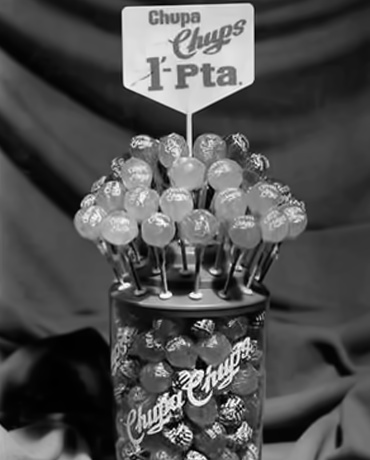}}
\subfigure
{\includegraphics[height=0.181\textwidth]{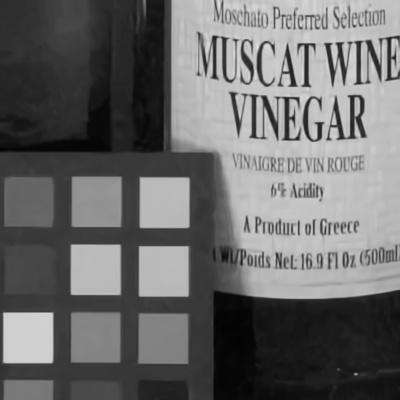}}
\subfigure
{\includegraphics[height=0.181\textwidth]{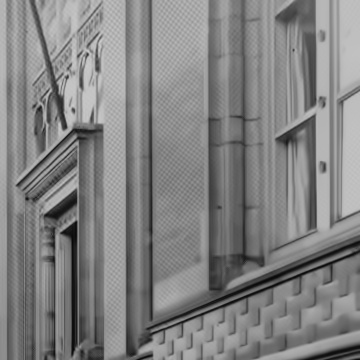}}
\subfigure
{\includegraphics[height=0.181\textwidth]{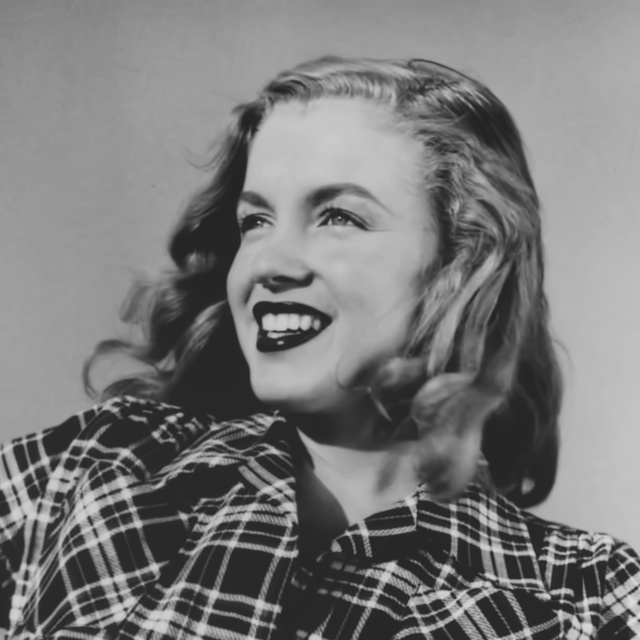}}

\subfigure
{\includegraphics[height=0.181\textwidth]{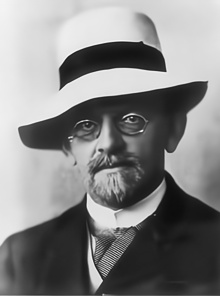}}
\subfigure
{\includegraphics[height=0.181\textwidth]{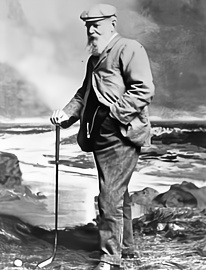}}
\subfigure
{\includegraphics[height=0.181\textwidth]{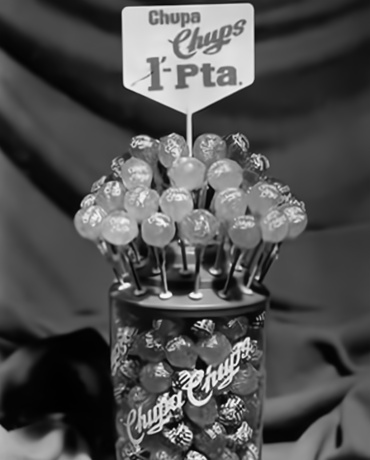}}
\subfigure
{\includegraphics[height=0.181\textwidth]{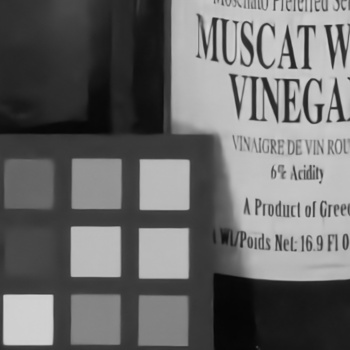}}
\subfigure
{\includegraphics[height=0.181\textwidth]{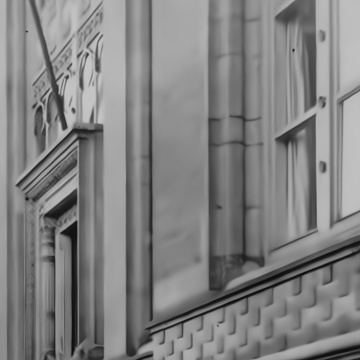}}
\subfigure
{\includegraphics[height=0.181\textwidth]{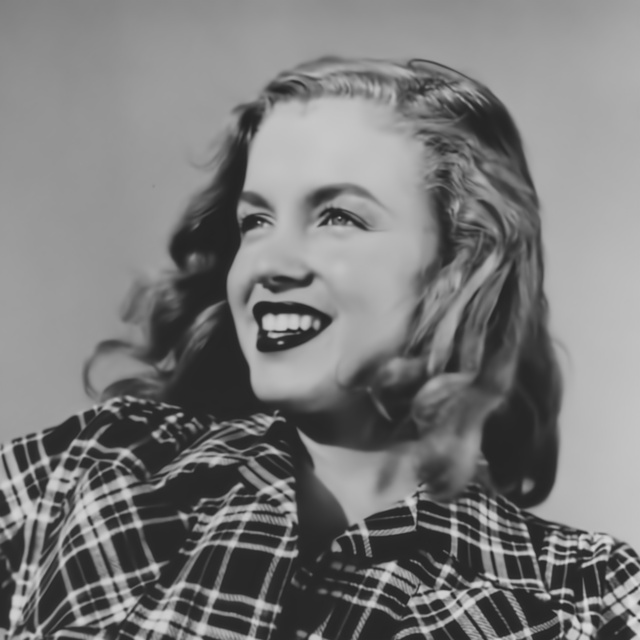}}

\subfigure
{\includegraphics[height=0.181\textwidth]{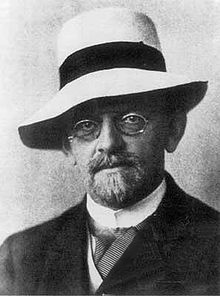}}
\subfigure
{\includegraphics[height=0.181\textwidth]{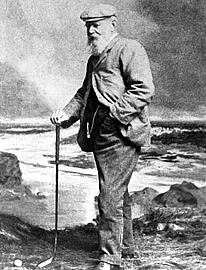}}
\subfigure
{\includegraphics[height=0.181\textwidth]{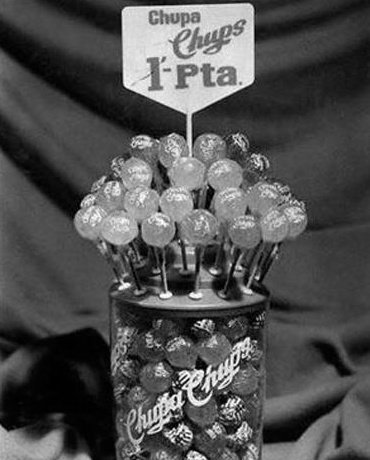}}
\subfigure
{\includegraphics[height=0.181\textwidth]{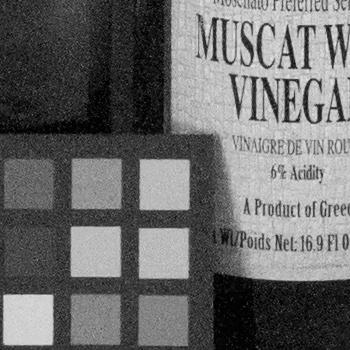}}
\subfigure
{\includegraphics[height=0.181\textwidth]{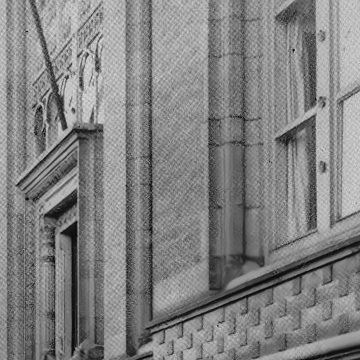}}
\subfigure
{\includegraphics[height=0.181\textwidth]{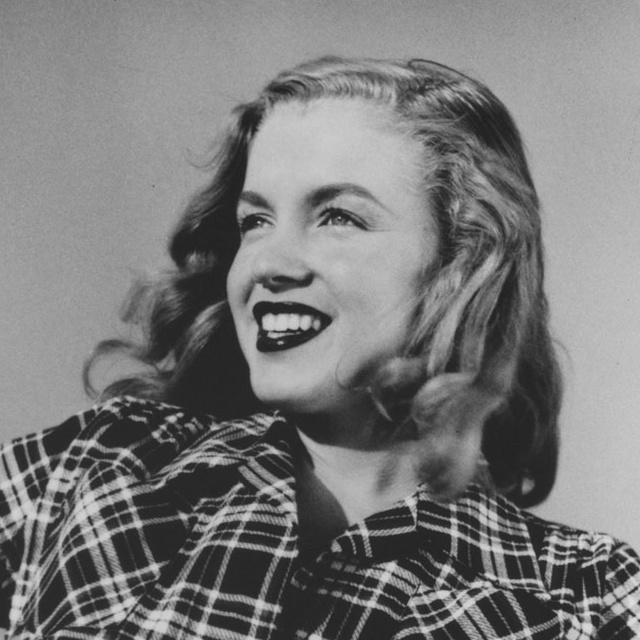}}

\subfigure[\emph{David Hilbert}]
{\includegraphics[height=0.181\textwidth]{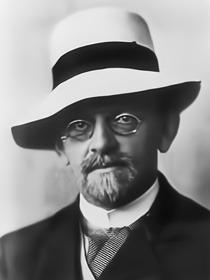}}
\subfigure[ \emph{Old Tom Morris}]
{\includegraphics[height=0.181\textwidth]{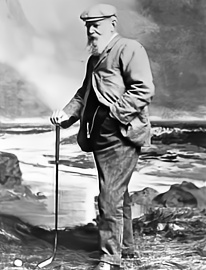}}
\subfigure[\emph{Chupa Chups}]
{\includegraphics[height=0.181\textwidth]{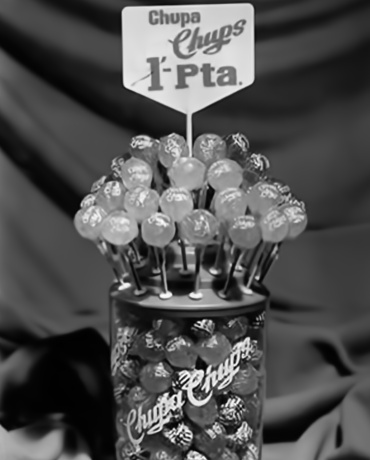}}
\subfigure[\emph{Vinegar}]
{\includegraphics[height=0.181\textwidth]{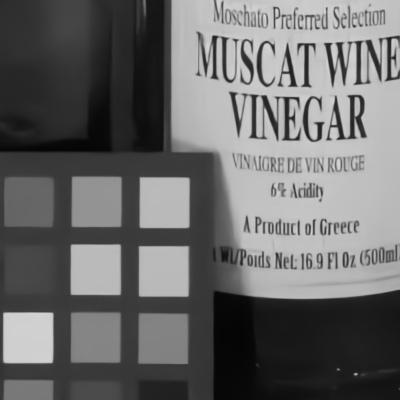}}
\subfigure[\emph{Building}]
{\includegraphics[height=0.181\textwidth]{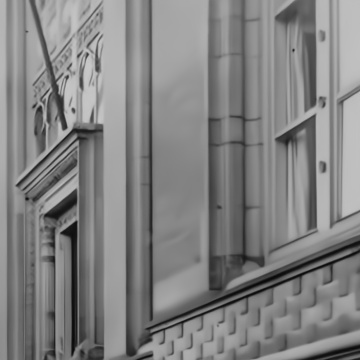}}
\subfigure[\emph{Marilyn}]
{\includegraphics[height=0.181\textwidth]{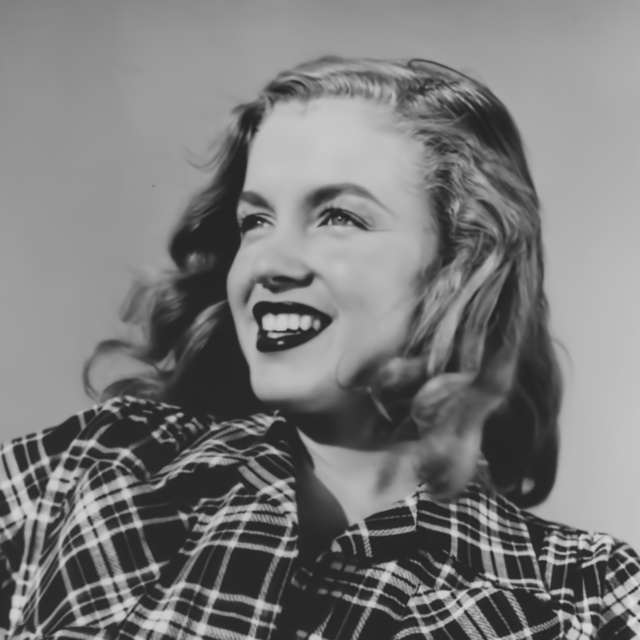}}
\caption{Grayscale image denoising results by different methods on real noisy images. From top to bottom: noisy images, denoised images by Noise Clinic, denoised images by BM3D, denoised images by DnCNN, denoised images by DnCNN-B, denoised images by FFDNet. (a) $\sigma$ = 14 (15 for DnCNN); (b) $\sigma$ = 15; (c) $\sigma$ = 10; (d) $\sigma$ = 20; (e) $\sigma$ = 20; (f) $\sigma$ = 7 (10 for DnCNN).}\label{fig_rn1}
\end{center}
\end{figure*}

Since there is no ground-truth image for a real noisy image, visual comparison is employed to evaluate the performance of FFDNet. We choose BM3D for comparison because it is widely accepted as a benchmark for denoising applications. Given a noisy image, the same input noise level is used for BM3D and FFDNet. Another CNN-based denoising method DnCNN and a blind denoising method Noise Clinic~\cite{lebrun2015noise} are also used for comparison. Note that, apart from the non-blind DnCNN models for specific noise levels, the blind DnCNN model (i.e., DnCNN-B) trained on noise level range of $[0, 55]$ is also used for grayscale image denoising. For color image denoising, the blind CDnCNN-B is used for comparison.

Fig.~\ref{fig_rn1} compares the grayscale image denoising results of Noise Clinic, BM3D, DnCNN, DnCNN-B and FFDNet on RNI6 images. As one can see, Noise Clinic reduces much the noise, but it also generates many algorithm-induced artifacts. BM3D, DnCNN and FFDNet produce more visually pleasant results.
While the non-blind DnCNN models perform favorably, the blind DnCNN-B model performs poorly in removing the non-AWGN real noise. This phenomenon clearly demonstrates the better generalization ability of non-blind model over blind one for controlling the trade-off between noise removal and detail preservation.
It is worth noting that, for image ``\emph{Building}'' which contains structured noise, Noise Clinic and BM3D fail to remove those structured noises since the structured noises fit the nonlocal self-similarity prior adopted in Noise Clinic and BM3D. In contrast, FFDNet and DnCNN successfully remove such noise without losing underlying image textures.

\begin{figure*}[!htbp]
\begin{center}
\subfigure
{\includegraphics[height=0.171\textwidth]{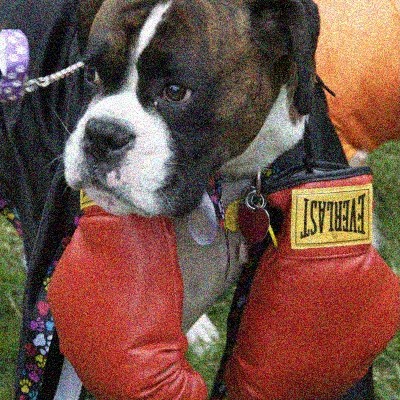}}
\subfigure
{\includegraphics[height=0.171\textwidth]{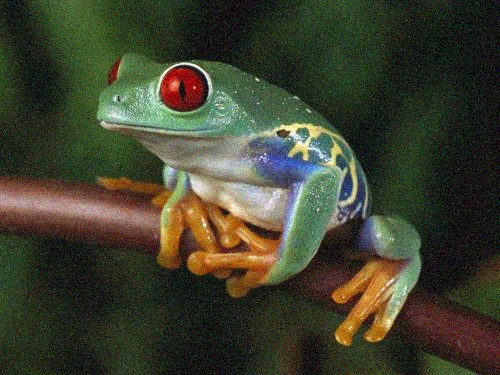}}
\subfigure
{\includegraphics[height=0.171\textwidth]{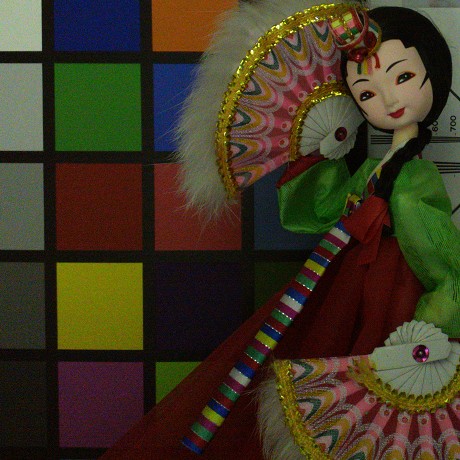}}
\subfigure
{\includegraphics[height=0.171\textwidth]{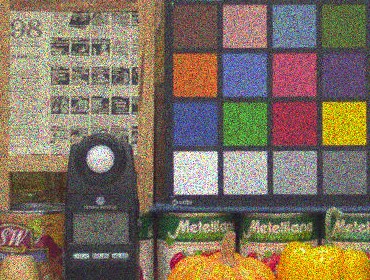}}
\subfigure
{\includegraphics[height=0.171\textwidth]{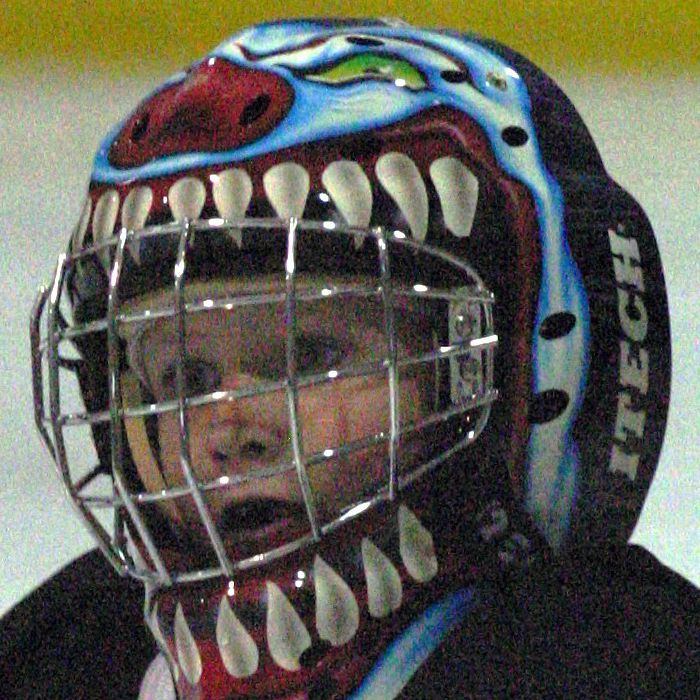}}

\subfigure
{\includegraphics[height=0.171\textwidth]{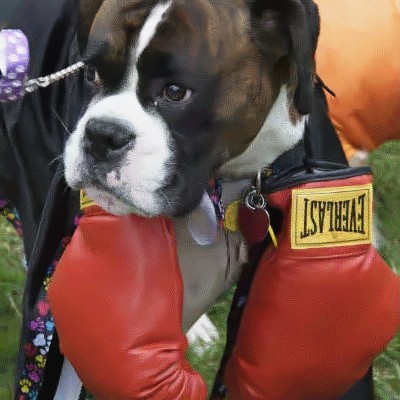}}
\subfigure
{\includegraphics[height=0.171\textwidth]{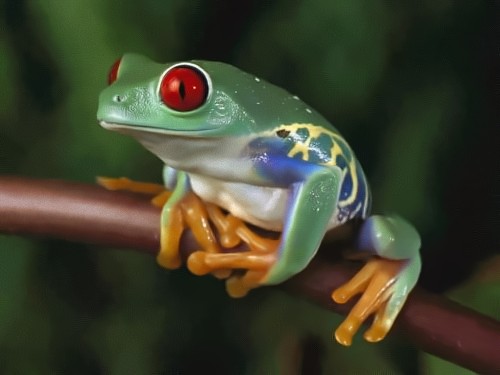}}
\subfigure
{\includegraphics[height=0.171\textwidth]{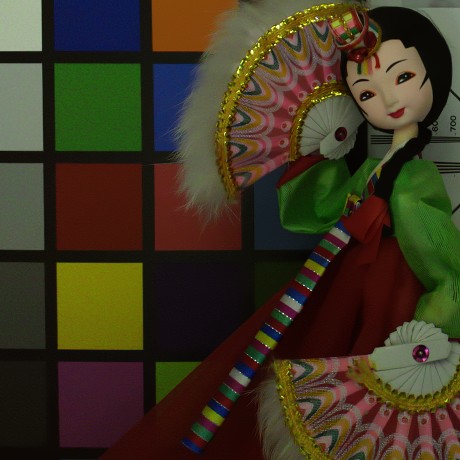}}
\subfigure
{\includegraphics[height=0.171\textwidth]{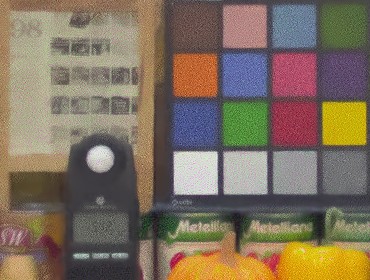}}
\subfigure
{\includegraphics[height=0.171\textwidth]{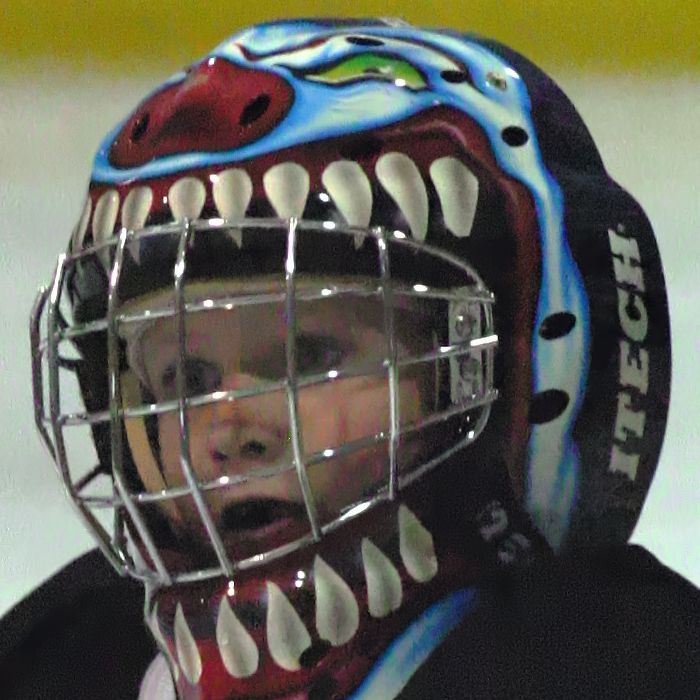}}

\subfigure
{\includegraphics[height=0.171\textwidth]{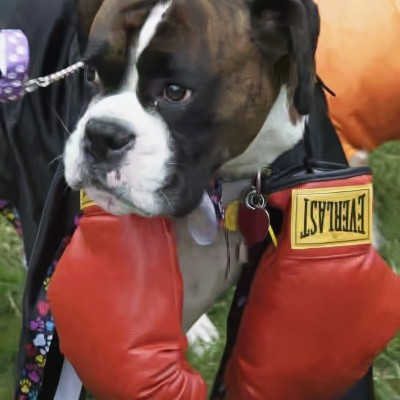}}
\subfigure
{\includegraphics[height=0.171\textwidth]{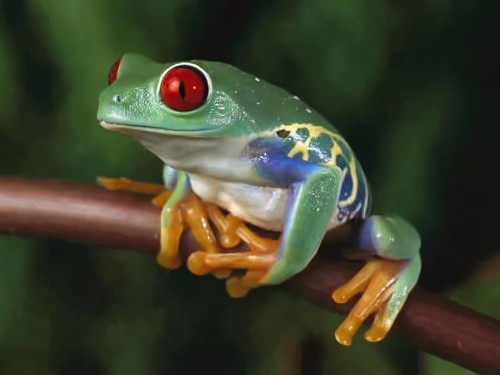}}
\subfigure
{\includegraphics[height=0.171\textwidth]{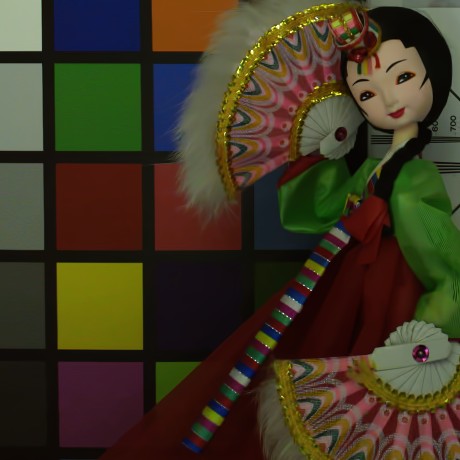}}
\subfigure
{\includegraphics[height=0.171\textwidth]{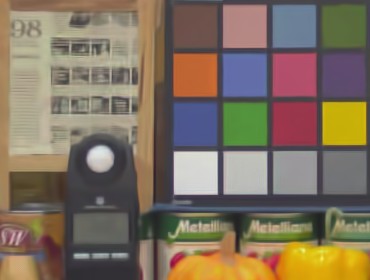}}
\subfigure
{\includegraphics[height=0.171\textwidth]{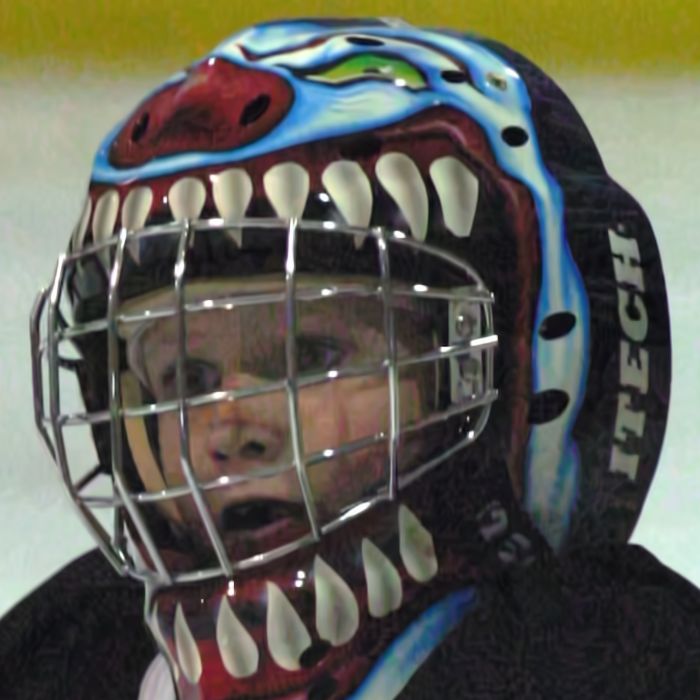}}

\subfigure
{\includegraphics[height=0.171\textwidth]{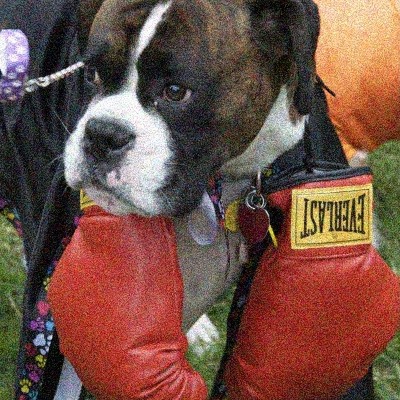}}
\subfigure
{\includegraphics[height=0.171\textwidth]{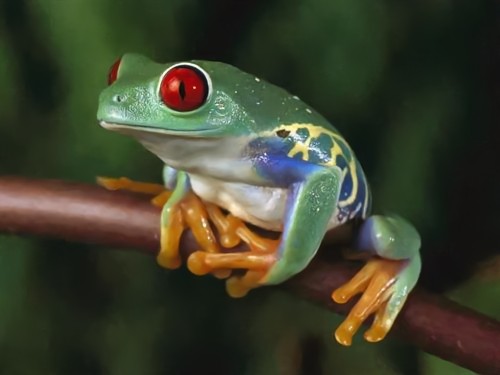}}
\subfigure
{\includegraphics[height=0.171\textwidth]{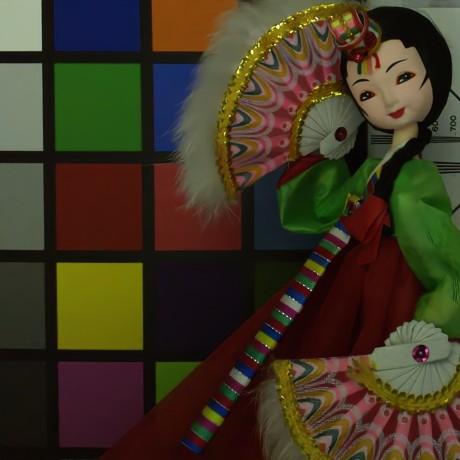}}
\subfigure
{\includegraphics[height=0.171\textwidth]{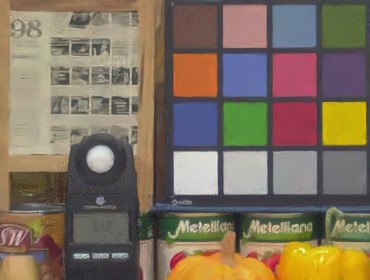}}
\subfigure
{\includegraphics[height=0.171\textwidth]{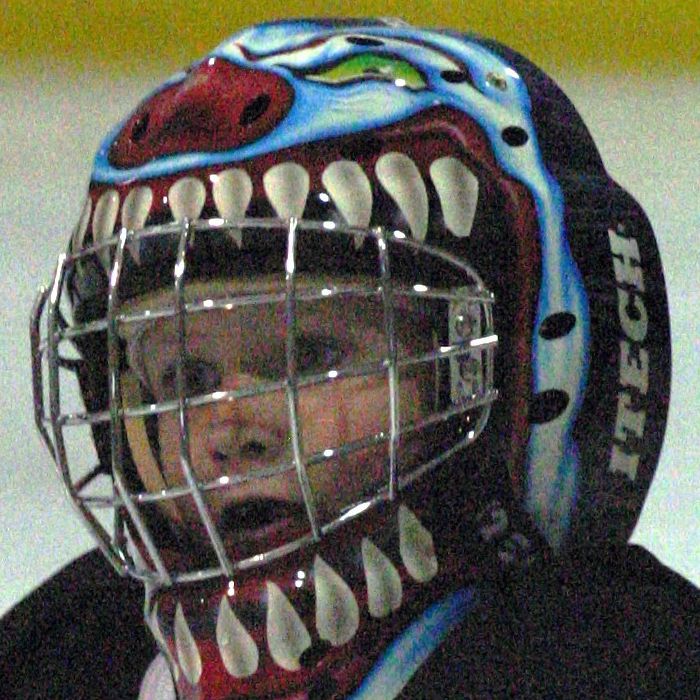}}\setcounter{subfigure}{0}

\subfigure[\emph{Dog}]
{\includegraphics[height=0.171\textwidth]{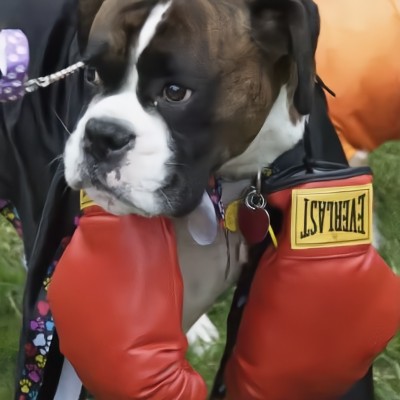}}
\subfigure[\emph{Frog}]
{\includegraphics[height=0.171\textwidth]{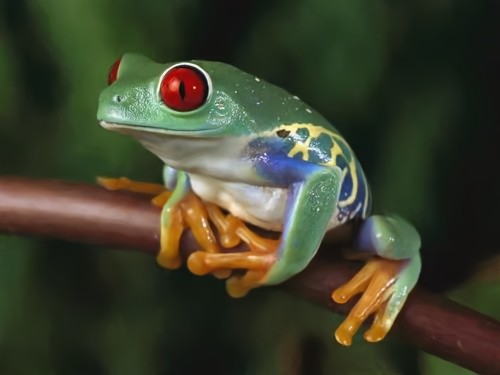}}
\subfigure[\emph{Pattern1}]
{\includegraphics[height=0.171\textwidth]{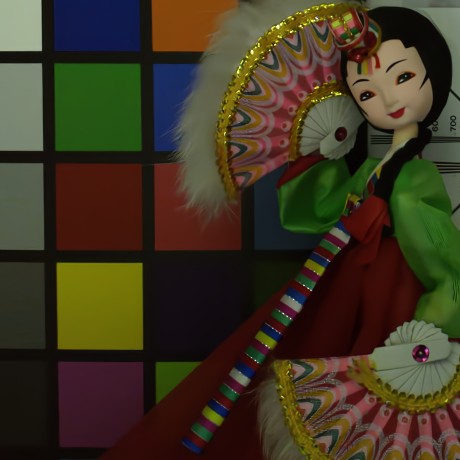}}
\subfigure[\emph{Pattern2}]
{\includegraphics[height=0.171\textwidth]{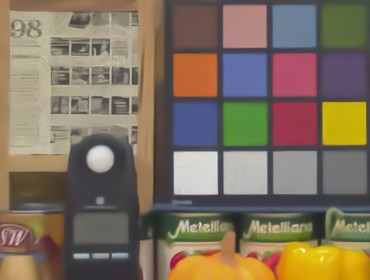}}
\subfigure[\emph{Boy}]
{\includegraphics[height=0.171\textwidth]{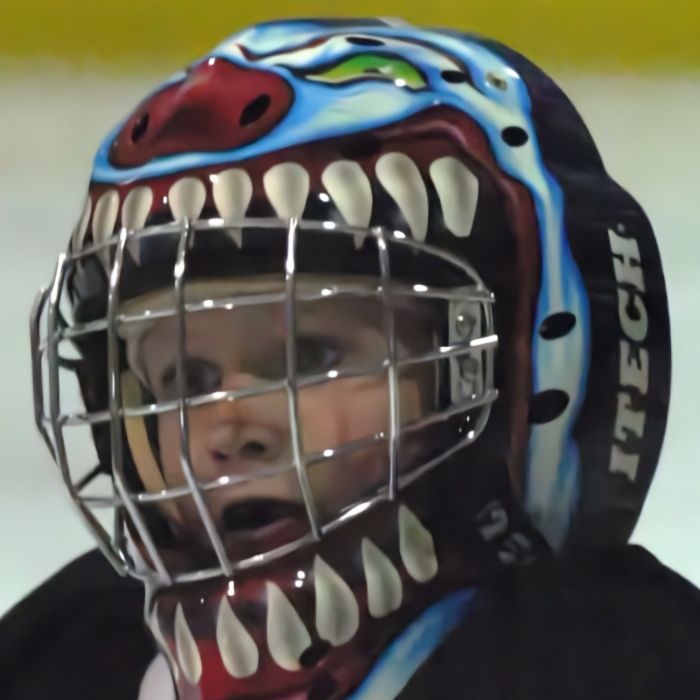}}
\caption{Color image denoising results by different methods on real noisy images. From top to bottom: noisy images, denoised images by Noise Clinic, denoised images by CBM3D, denoised images by CDnCNN-B, denoised images by FFDNet. (a) $\sigma$ = 28; (b) $\sigma$ = 15; (c) $\sigma$ = 12; (d) $\sigma$ = 40; (e) $\sigma$ = 45.}\label{fig_rn2}
\end{center}
\end{figure*}

Fig.~\ref{fig_rn2} shows the denoising results of Noise Clinic, CBM3D, {CDnCNN-B} and FFDNet on five color noisy images from RNI15.
It can be seen that CDnCNN-B yields very pleasing results for noisy image with AWGN-like noise such as image ``\emph{Frog}'', and is still unable to handle non-AWGN noise.
Notably, from the denoising results of ``\emph{Boy}'', one can see that CBM3D remains the structured color noise unremoved whereas FFDNet removes successfully such kind of noise. We can conclude that while the nonlocal self-similarity prior helps to remove random noise, it hinders the removal of structured noise. In comparison, the prior implicitly learned by CNN is able to remove both random noise and structured noise.

\begin{figure*}[!htbp]
\begin{center}
\subfigure
{\includegraphics[height=0.169\textwidth]{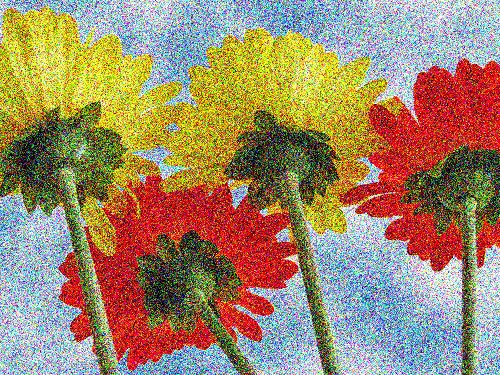}}
\subfigure
{\includegraphics[height=0.169\textwidth]{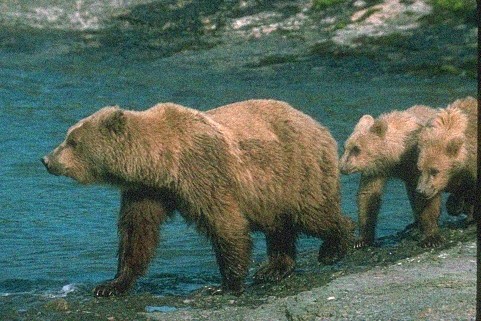}}
\subfigure
{\includegraphics[height=0.169\textwidth]{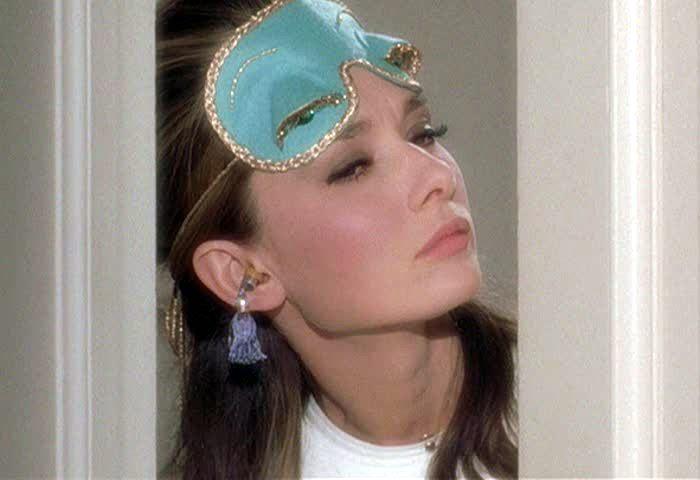}}
\subfigure
{\includegraphics[height=0.169\textwidth]{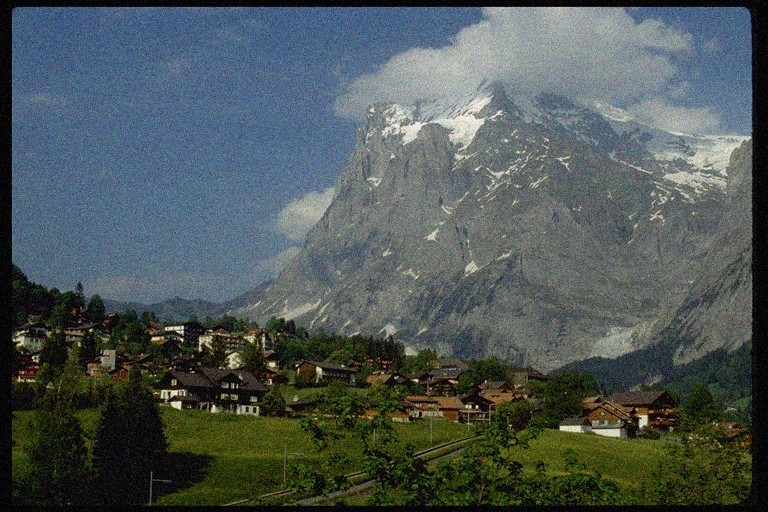}}\setcounter{subfigure}{0}
\subfigure[\emph{Flowers}]
{\includegraphics[height=0.169\textwidth]{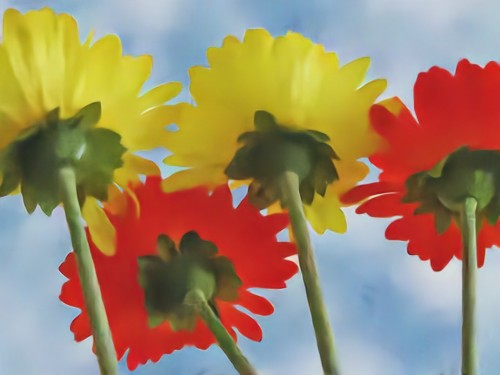}}
\subfigure[\emph{Bears}]
{\includegraphics[height=0.169\textwidth]{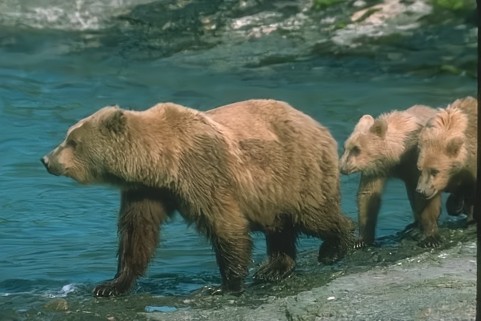}}
\subfigure[\emph{Audrey Hepburn}]
{\includegraphics[height=0.169\textwidth]{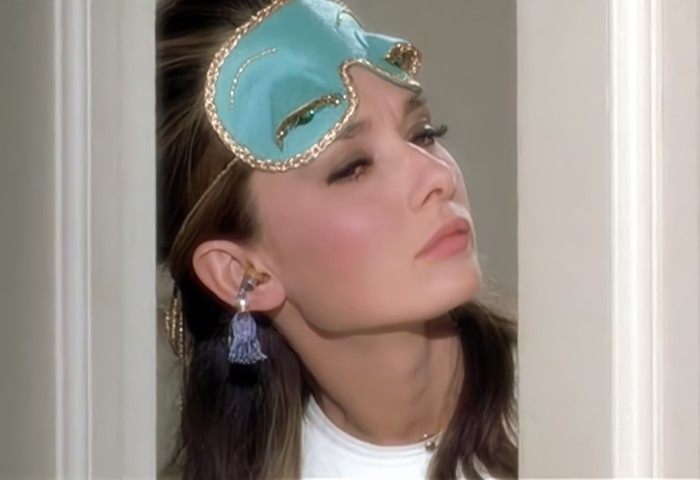}}
\subfigure[\emph{Postcards}]
{\includegraphics[height=0.169\textwidth]{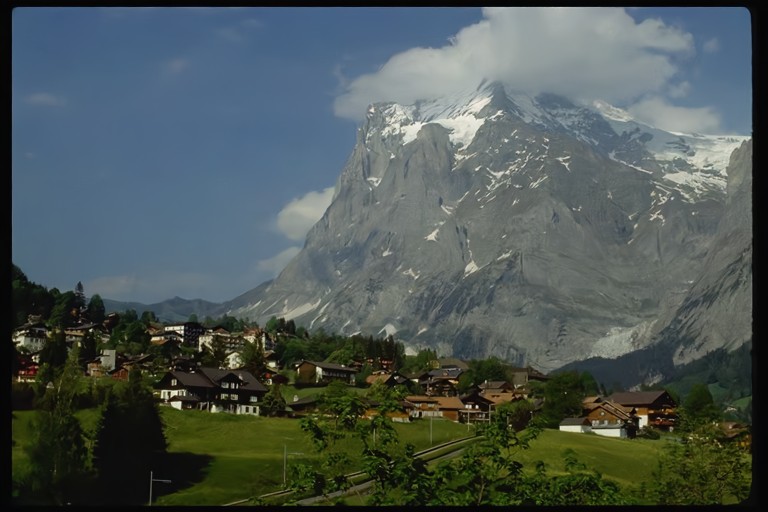}}

\subfigure
{\includegraphics[height=0.21\textwidth]{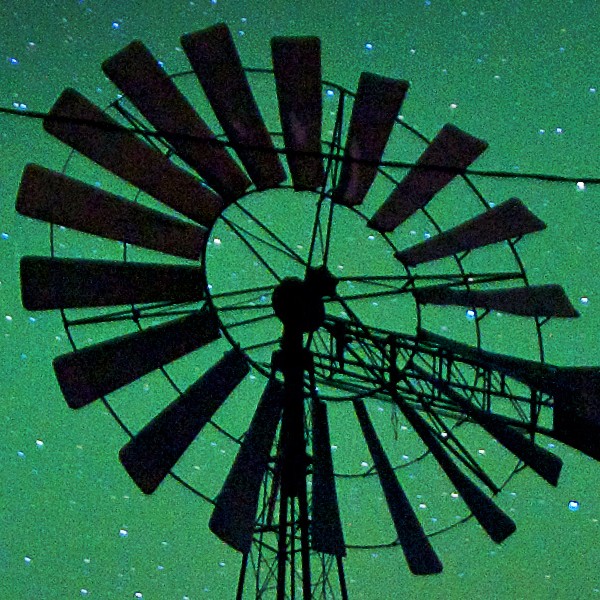}}
\subfigure
{\includegraphics[height=0.21\textwidth]{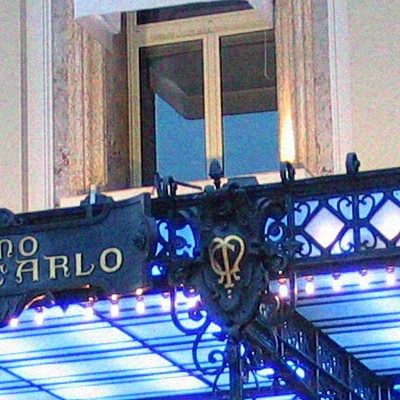}}
\subfigure
{\includegraphics[height=0.21\textwidth]{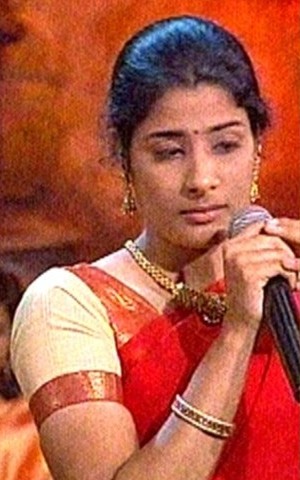}}
\subfigure
{\includegraphics[height=0.21\textwidth]{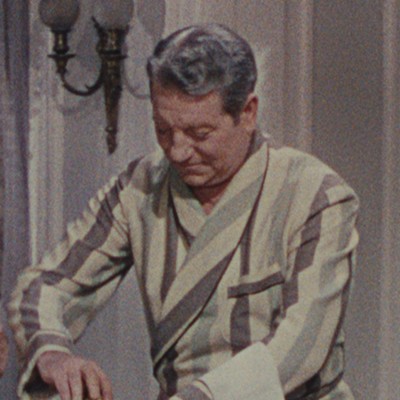}}
\subfigure
{\includegraphics[height=0.21\textwidth]{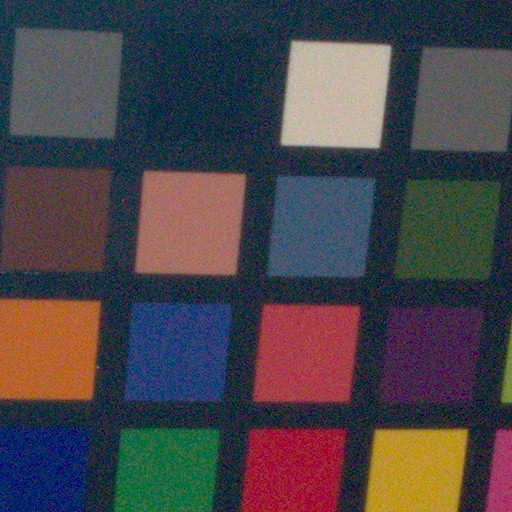}}\setcounter{subfigure}{4}
\subfigure[\emph{Stars}]
{\includegraphics[height=0.21\textwidth]{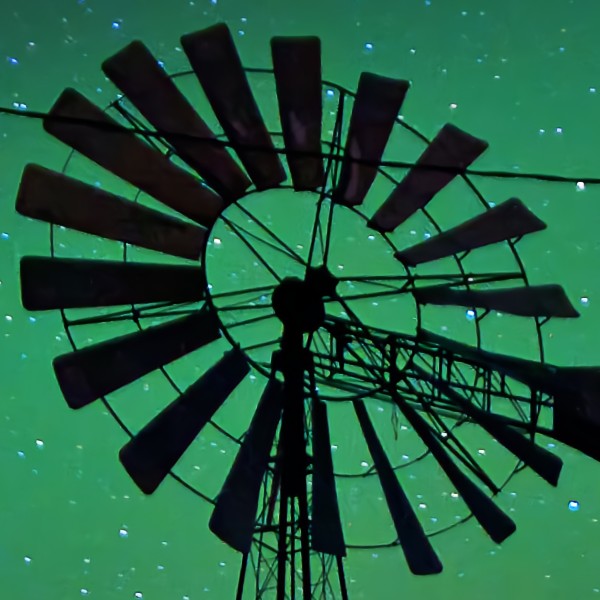}}
\subfigure[\emph{Window}]
{\includegraphics[height=0.21\textwidth]{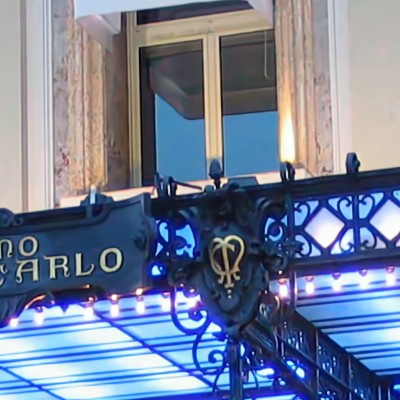}}
\subfigure[\emph{Singer}]
{\includegraphics[height=0.21\textwidth]{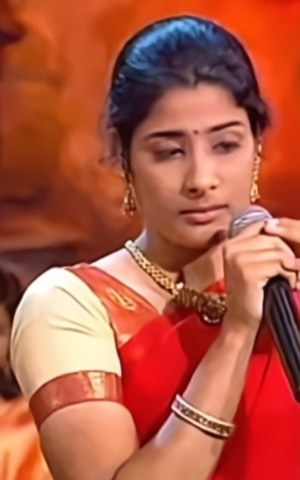}}
\subfigure[\emph{Movie}]
{\includegraphics[height=0.21\textwidth]{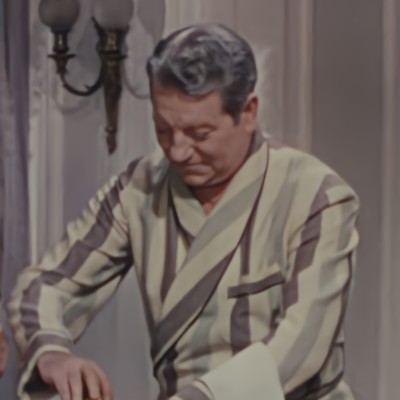}}
\subfigure[\emph{Pattern3}]
{\includegraphics[height=0.21\textwidth]{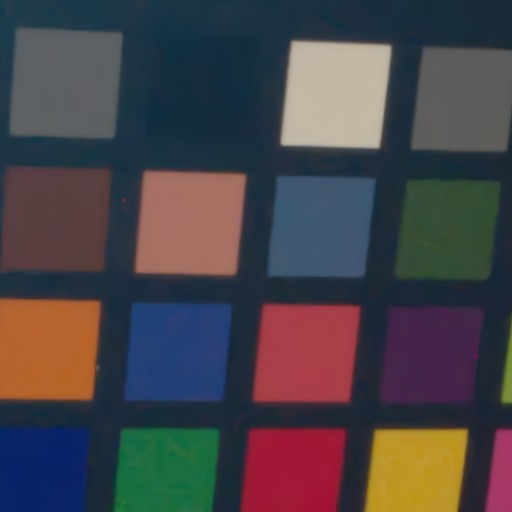}}
\caption{More denoising results of FFDNet on real image denoising. (a) $\sigma$ = 70; (b) $\sigma$ = 15; (c) $\sigma$ = 10; (d) $\sigma$ = 15; (e) $\sigma$ = 18; (f) $\sigma$ = 15; (g) $\sigma$ = 30;  (h) $\sigma$ = 12; (i) $\sigma$ = 25.}\label{fig_rn3}
\end{center}
\end{figure*}

Fig.~\ref{fig_rn3} further shows more visual results of FFDNet on the other nine images from RNI15. It can be seen that  FFDNet can handle various kinds of noises, such as JPEG lossy compression noise (see image ``\emph{Audrey Hepburn}''), and video noise (see image ``\emph{Movie}'').

\begin{figure*}[!htbp]
\begin{center}
\subfigure[]
{\includegraphics[width=0.161\textwidth]{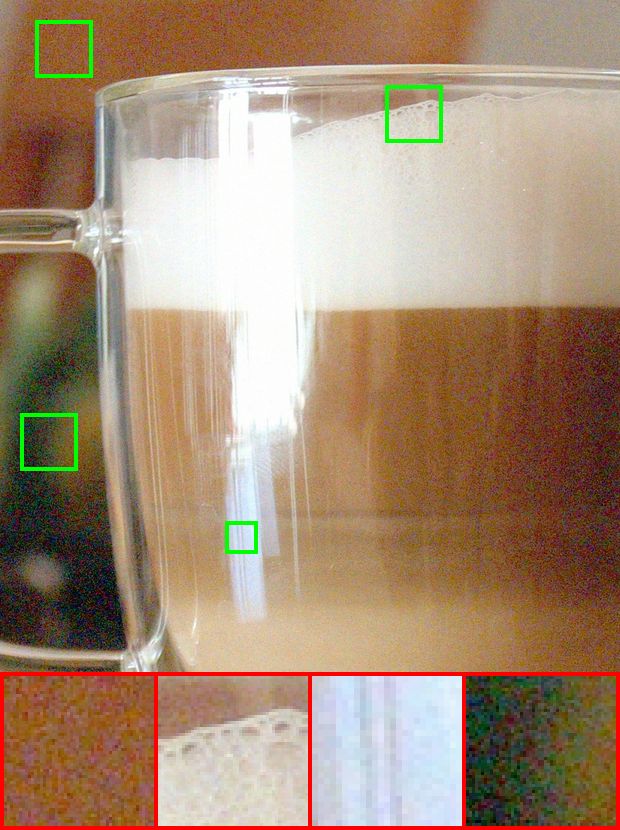}}
\subfigure[]
{\includegraphics[width=0.161\textwidth]{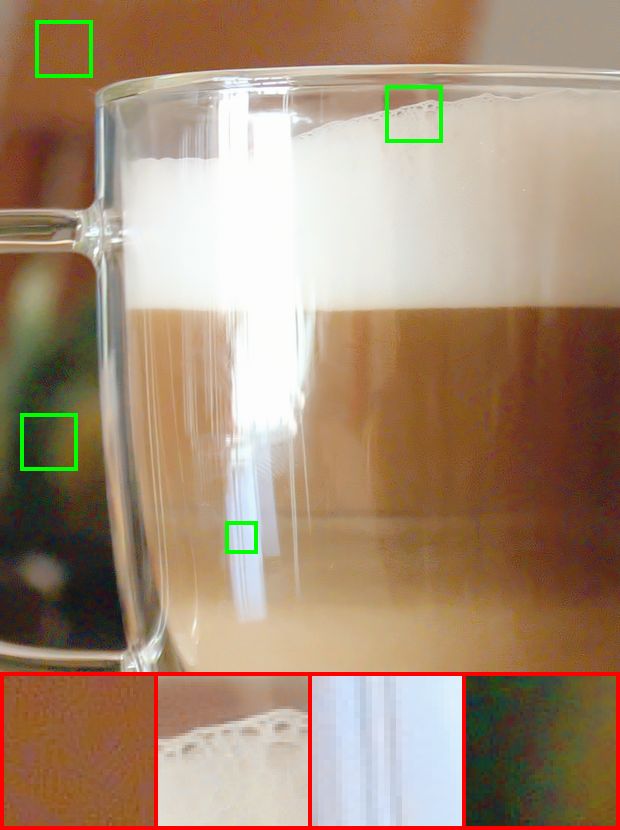}}
\subfigure[]
{\includegraphics[width=0.161\textwidth]{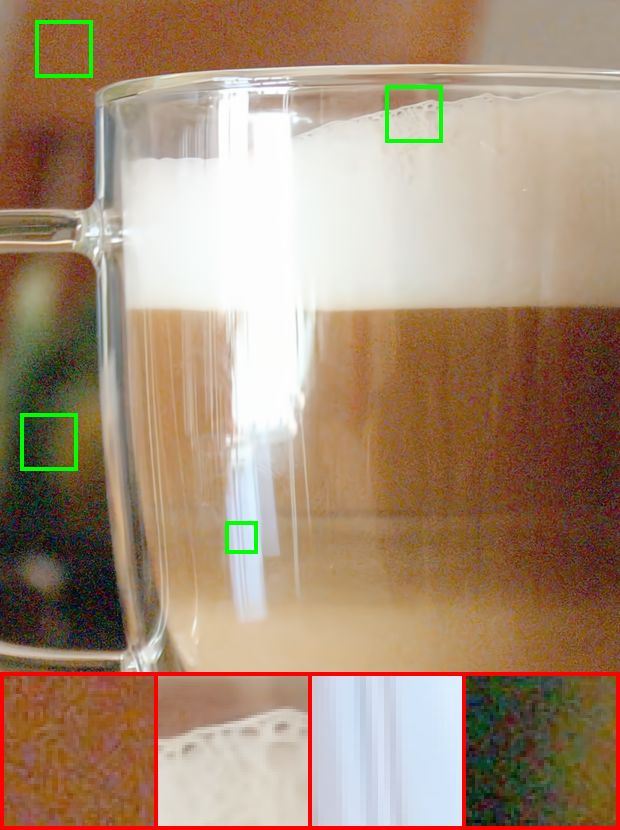}}
\subfigure[]
{\includegraphics[width=0.161\textwidth]{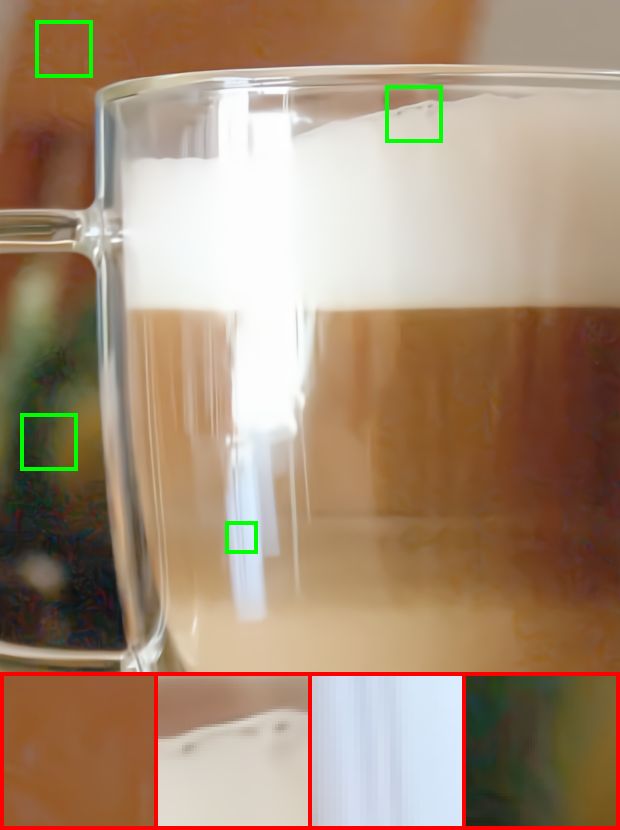}}
\subfigure[]
{\includegraphics[width=0.161\textwidth]{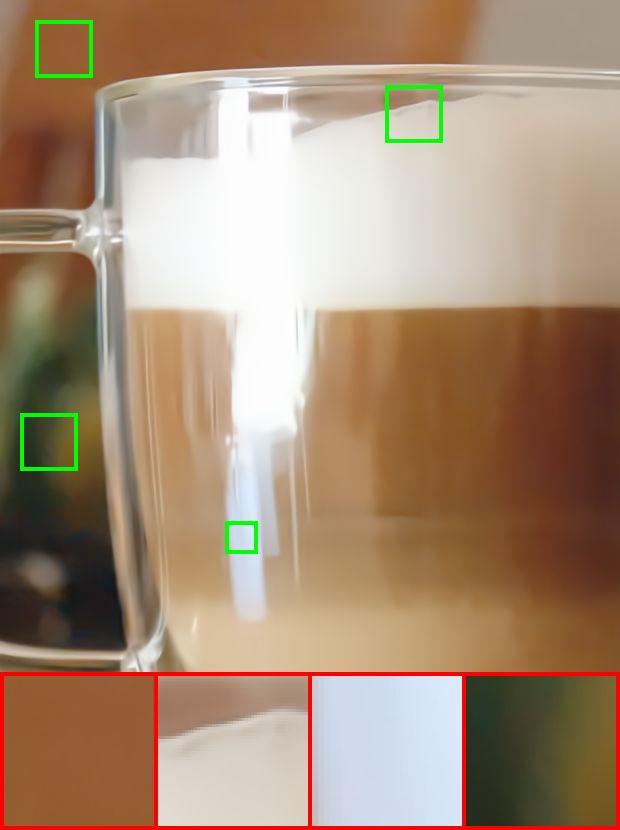}}
\subfigure[]
{\includegraphics[width=0.161\textwidth]{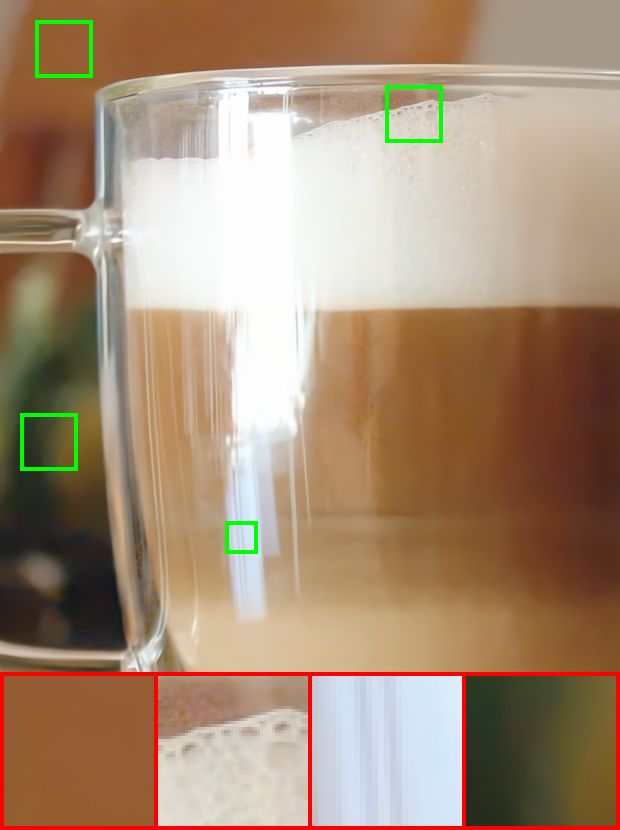}}
\caption{An example of FFDNet on image ``\emph{Glass}'' with spatially variant noise. (a) Noisy image; (b) Denoised image by Noise Clinic; (c) Denoised image by FFDNet with $\sigma$ = 10; (d) Denoised image by FFDNet with $\sigma$ = 25; (e) Denoised image by FFDNet with $\sigma$ = 35; (f) Denoised image by FFDNet with non-uniform noise level map.}\label{fig_rn4}
\end{center}
\end{figure*}

Fig.~\ref{fig_rn4} shows a more challenging example to demonstrate the advantage of FFDNet for denoising noisy images with spatially variant noise.
We select five typical regions to estimate the noise levels, including two background regions, the coffee region, the milk-foam region, and the specular reflection region. In our experiment, we manually and interactively set $\sigma$ = 10 for the milk-foam and specular reflection regions, $\sigma$ = 35 for the background region with high noise (i.e., green region), and $\sigma$ = 25 for the other regions. We then interpolate the non-uniform noise level map for the whole image based on the estimated five noise levels.
As one can see, while FFDNet with a small uniform input noise level can recover the details of regions with low noise level, it fails to remove strong noise. On the other hand, FFDNet with a large uniform input noise level can remove strong noise but it will also smooth out the details in the region with low noise level. In contrast, the denoising result with a proper non-uniform noise level map not only preserves image details but also removes the strong noise.

Finally, according to the above experiments on real noisy images, we can see that the FFDNet model trained with un-quantized image data performs well on 8-bit quantized real noisy images.

\subsection{Running Time}\label{section_runtime}

Table~\ref{table4} lists the running time results of BM3D, DnCNN and FFDNet for denoising grayscale and color images with size 256$\times$256, 512$\times$512 and 1,024$\times$1,024.
The evaluation was performed in Matlab (R2015b) environment on a computer with a six-core Intel(R) Core(TM) i7-5820K CPU @ 3.3GHz, 32 GB of RAM and an Nvidia Titan X Pascal GPU.
For BM3D, we evaluate its running time by denoising images with noise level 25. For DnCNN, the grayscale and color image denoising models have 17 and 20 convolution layers, respectively.
The Nvidia cuDNN-v5.1 deep learning library is used to accelerate the computation of DnCNN and FFDNet. The memory transfer time between CPU and GPU is also counted. Note that DnCNN and FFDNet can be implemented
with both single-threaded (ST) and multi-threaded (MT) CPU computations.

\begin{table}[!htbp]\footnotesize\arrayrulewidth0.5pt
\caption{Running time (in seconds) of different methods for denoising images with size 256$\times$256, 512$\times$512 and 1,024$\times$1,024}
\center
\begin{tabular}{|p{0.9cm}<{\centering}|p{1cm}<{\centering}|p{0.56cm}<{\centering}|p{0.56cm}<{\centering}|p{0.56cm}<{\centering}|p{0.56cm}<{\centering}|p{0.56cm}<{\centering}|p{0.56cm}<{\centering}|}
  \hline
  \multirow{2}{*}{Methods} & \multirow{2}{*}{Device} & \multicolumn{2}{c|}{\scriptsize 256$\times$256}   & \multicolumn{2}{c|}{\scriptsize 512$\times$512}  & \multicolumn{2}{c|}{\scriptsize 1,024$\times$1,024}  \\ 
        &   & \cellcolor[rgb]{.9,.9,.9} Gray & \cellcolor[rgb]{.75,.75,.75}Color   & \cellcolor[rgb]{.9,.9,.9}Gray  & \cellcolor[rgb]{.75,.75,.75}Color  & \cellcolor[rgb]{.9,.9,.9}Gray & \cellcolor[rgb]{.75,.75,.75}Color  \\ \hline
  BM3D& CPU(ST)  & \cellcolor[rgb]{.9,.9,.9}0.59 & \cellcolor[rgb]{.75,.75,.75}0.98 & \cellcolor[rgb]{.9,.9,.9}2.52 & \cellcolor[rgb]{.75,.75,.75}3.57  & \cellcolor[rgb]{.9,.9,.9}10.77 & \cellcolor[rgb]{.75,.75,.75}20.15 \\\hline
                                        &  CPU(ST)  & \cellcolor[rgb]{.9,.9,.9}2.14  & \cellcolor[rgb]{.75,.75,.75}2.44 & \cellcolor[rgb]{.9,.9,.9}8.63 & \cellcolor[rgb]{.75,.75,.75}9.85 & \cellcolor[rgb]{.9,.9,.9}32.82& \cellcolor[rgb]{.75,.75,.75}38.11 \\
   DnCNN & CPU(MT)  & \cellcolor[rgb]{.9,.9,.9}0.74& \cellcolor[rgb]{.75,.75,.75}0.98 &\cellcolor[rgb]{.9,.9,.9}3.41 & \cellcolor[rgb]{.75,.75,.75}4.10 & \cellcolor[rgb]{.9,.9,.9}12.10 & \cellcolor[rgb]{.75,.75,.75}15.48 \\
                                         &  GPU  & \cellcolor[rgb]{.9,.9,.9}0.011 & \cellcolor[rgb]{.75,.75,.75}0.014 &  \cellcolor[rgb]{.9,.9,.9}0.033& \cellcolor[rgb]{.75,.75,.75}0.040 & \cellcolor[rgb]{.9,.9,.9}0.124& \cellcolor[rgb]{.75,.75,.75}0.167  \\\hline
           & CPU(ST)  & \cellcolor[rgb]{.9,.9,.9}0.44& \cellcolor[rgb]{.75,.75,.75}0.62 & \cellcolor[rgb]{.9,.9,.9}1.81 & \cellcolor[rgb]{.75,.75,.75}2.51 &  \cellcolor[rgb]{.9,.9,.9}7.24 & \cellcolor[rgb]{.75,.75,.75}10.17 \\
                     FFDNet                    &  CPU(MT)  & \cellcolor[rgb]{.9,.9,.9}0.18  &\cellcolor[rgb]{.75,.75,.75}0.21 & \cellcolor[rgb]{.9,.9,.9}0.73 & \cellcolor[rgb]{.75,.75,.75}0.98 &  \cellcolor[rgb]{.9,.9,.9}2.96 & \cellcolor[rgb]{.75,.75,.75}3.95\\
                                         &  GPU  & \cellcolor[rgb]{.9,.9,.9}0.006  &\cellcolor[rgb]{.75,.75,.75}0.008 & \cellcolor[rgb]{.9,.9,.9}0.012 & \cellcolor[rgb]{.75,.75,.75}0.017 &  \cellcolor[rgb]{.9,.9,.9}0.038 & \cellcolor[rgb]{.75,.75,.75}0.057\\
  \hline
\end{tabular}
\label{table4}
\end{table}

From Table~\ref{table4}, we have the following observations. First, BM3D spends much more time on denoising color images than grayscale images. The reason is that, compared to gray-BM3D, CBM3D needs extra time to denoise the chrominance components after luminance-chrominance color transformation. Second, while DnCNN can benefit from GPU computation for fast implementation, it has comparable CPU time to BM3D. Third, FFDNet spends almost the same time for processing grayscale and color images. More specifically, FFDNet with multi-threaded implementation is about three times faster than DnCNN and BM3D on CPU, and much faster than DnCNN on GPU. Even with single-threaded implementation, FFDNet is also faster than BM3D. Taking denoising performance and flexibility into consideration, FFDNet is very competitive for practical applications.

\section{Conclusion}
\label{sec:conclusion}
In this paper, we proposed a new CNN model, namely FFDNet, for fast, effective and flexible discriminative denoising. To achieve this goal, several techniques were utilized in network design and training, such as the use of noise level map as input and denoising in downsampled sub-images space. The results on synthetic images with AWGN demonstrated that FFDNet can not only produce state-of-the-art results when input noise level matches ground-truth noise level, {but also have the ability to robustly control
the trade-off between noise reduction and detail preservation}. The results on images with spatially variant AWGN validated the flexibility of FFDNet for handing inhomogeneous noise. The results on real noisy images further demonstrated that FFDNet can deliver perceptually appealing denoising results. Finally, the running time comparisons showed the faster speed of FFDNet over other competing methods such as BM3D. Considering its flexibility, efficiency and effectiveness, FFDNet provides a practical solution to CNN denoising applications.

\bibliographystyle{IEEEtran}
\bibliography{IEEEabrv,egbib}

\end{document}